\pdfoutput=1
\documentclass[11pt]{article}

\usepackage[final]{acl}
\usepackage{algorithm}
\usepackage{algpseudocode}
\usepackage{times}
\usepackage{latexsym}
\usepackage{graphicx}
\usepackage{amsmath} 
\usepackage{booktabs}
\usepackage{hyperref}
\usepackage{url}
\usepackage{amssymb} 
\usepackage{xcolor}
\usepackage{hyperref}       % hyperlinks
\usepackage{url}            % simple URL typesetting
\usepackage{booktabs}       % professional-quality tables
\usepackage{amsfonts}       % blackboard math symbols
\usepackage{nicefrac}       % compact symbols for 1/2, etc.
\usepackage{array}
\usepackage{booktabs}
\usepackage{xcolor}
\usepackage{colortbl}
\usepackage{graphicx}
\usepackage{multirow}
\usepackage[table]{xcolor}
\usepackage{booktabs}
\usepackage{longtable}
\usepackage{todonotes}
\usepackage{graphicx}
\usepackage{subcaption}
\usepackage[font=small,labelfont=bf]{caption}
\usepackage[T1]{fontenc}
\usepackage[utf8]{inputenc}

\usepackage{microtype}

\usepackage{inconsolata}

\usepackage{graphicx}
\usepackage[normalem]{ulem}  % Add this in the preamble

\title{Scaling Vision–Language Models for Pharmaceutical Long-Form Video Reasoning on Industrial GenAI Platform}

\author{
  \textbf{Suyash Mishra\textsuperscript{a}},  
  \textbf{Qiang Li\textsuperscript{b}},
  \textbf{Satyanarayan Pati\textsuperscript{c}},
  \textbf{Srikanth Patil\textsuperscript{c}},
  \textbf{Narendra Baddu \textsuperscript{c}}
\\ 
  \textsuperscript{a}Roche,
  \textsuperscript{b}Accenture,  
  \textsuperscript{c}Involead,
\\
  \small{
    \textbf{Correspondence:}  
    \href{mailto:suyash.mishra@roche.com}{suyash.mishra@roche.com}},
    \href{mailto:qiang.i.li@accenture.com}{qiang.i.li@accenture.com}
    \href{mailto:satyanarayan.pati@involead.com}{satyanarayan.pati@involead.com}
\\ 
\small{ 
    \href{mailto:srikanth.patil@involead.com}{srikanth.patil@involead.com},
        \href{mailto:baddu.narendra@involead.com}{baddu.narendra@involead.com}}
    \\  
 }

\begin{document}
\maketitle

\begin{abstract}
Vision--Language Models (VLMs) have shown strong performance on multimodal reasoning tasks, yet most evaluations focus on short videos and assume unconstrained computational resources. In industrial settings such as pharmaceutical content understanding, practitioners must process long-form videos under strict GPU, latency, and cost constraints, where many existing approaches fail to scale. In this work, we present an industrial GenAI framework that processes over 200{,}000 PDFs, 25{,}326 videos across eight formats (e.g., MP4, M4V, etc.), and 888 multilingual audio files in more than 20 languages. Our study makes three contributions: (i) an industrial large-scale architecture for multimodal reasoning in pharmaceutical domains; (ii) empirical analysis of over 40 VLMs on two leading benchmarks (Video-MME and MMBench) and proprietary dataset of 25{,}326 videos across 14 disease areas; and (iii) four findings relevant to long-form video reasoning: the role of multimodality, attention mechanism trade-offs, temporal reasoning limits, and challenges of video splitting under GPU constraints. Results show 3--8$\times$ efficiency gains with SDPA attention on commodity GPUs, multimodality improving up to 8/12 task domains (especially length-dependent tasks), and clear bottlenecks in temporal alignment and keyframe detection across open- and closed-source VLMs.  Rather than proposing a new "A+B" model, this paper characterizes practical limits, trade-offs, and failure patterns of current VLMs under realistic deployment constraints, and provide actionable guidance for both researchers and practitioners designing scalable multimodal systems for long-form video understanding in industrial domains.

%Our work reframes VLM evaluation in industrial settings and also offers practical, scalable business solutions for multimodal reasoning under real-world GPU constraints, ultimately facilitating video derivative creation.

$\overline{\text{{* Patent application submitted to the EPO}}}$

\end{abstract}

\begin{figure*}[ht]
    \centering
    \begin{minipage}{1\textwidth}
        \centering
        \includegraphics[width=1\textwidth]{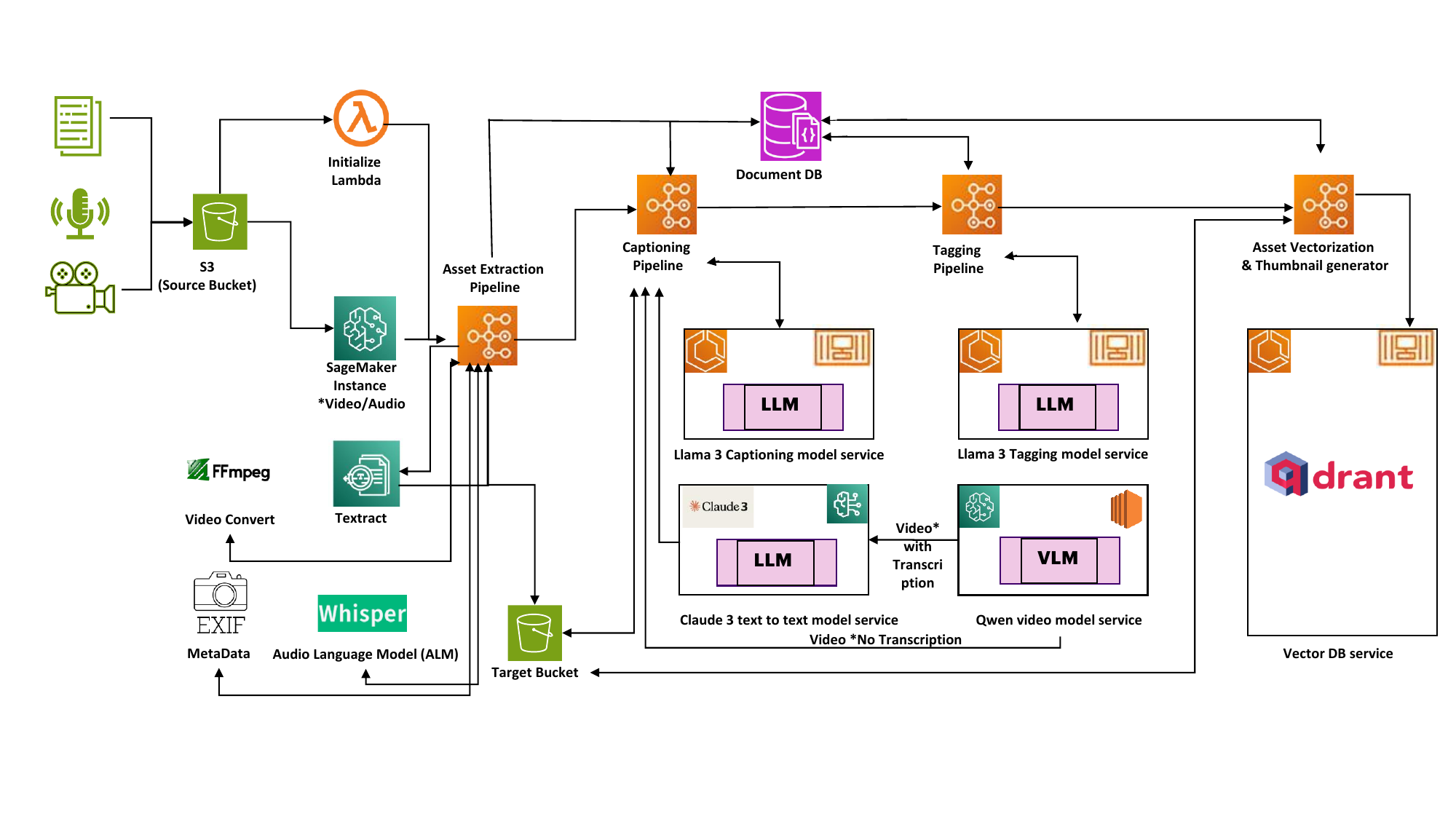} 
    \end{minipage}%
    \vspace{-40pt}
    \caption{System architecture of our GenAI platform for Natural Language (NL) search integrating LLMs, ALMs, and VLMs. The platform processed 25,326 videos, 888 audios covering > 20 languages.}
    \label{fig: architecture}
\end{figure*}

\section{Introduction}

%Large Language Models (LLMs) such as GPT-4 \citep{instructGPT} and LaMDA \citep{bard} have substantially enhanced public access to complex information, particularly in sectors such as healthcare and public services \citep{li-etal-2025-llms} . These models are celebrated for their capability to demystify intricate information, assisting in tasks ranging from routine inquiries to aiding clinical decision-making \citep{LLM}. 

%Now powered by Generative AI, especially VLM and ALM, multimodal system improves document consumption by integrating tables, graphs, charts, infographics, text, and video and audio \citep{fu2024video}. It analyzes content to uncover traceable insights such as facts and claims, and generates new material that complies with brand, expert domain, country, and global guidelines through rigorous quality checks. These Generative AI solutions \citep{vaswani2017attention, brown2020language, ramesh2021dalle, rombach2022stable} enable more efficient content creation and streamlined operations.

Large Language Models (LLMs) such as GPT-4 \citep{instructGPT} and LaMDA \citep{bard} have significantly improved access to complex information in domains including healthcare and public services (H\&PS) \citep{li-etal-2025-llms}. While these models excel at text-based reasoning, industrial use cases increasingly involve multimodal content spanning tables, graphs, charts, text, video and audio \citep{fu2024video}. In pharmaceutical industry, such assets include clinical trial recordings, conference lectures, promotional materials, and multilingual patient educational videos. Manual perception of these resources is inconsistent, costly, and infeasible at scale, particularly given compliance \& computation constraints.

Practical use cases including chat agents \citep{openai2023chatgpt}, briefing agencies \citep{openai2023chatgptintel}, document searching \citep{lewis2021retrievalaugmentedgenerationknowledgeintensivenlp}, text/video summarization \citep{li2025benchmarkevaluationsapplicationschallenges}, and document quality checks \citep{yang2024modifyingaienhancingessays} deliver transformative impacts on industry. %These applications streamline current processes, open up new opportunities.

Current research has emphasized novel architectures or benchmark leaderboards \citep{KnowledgeLLM}, but less attention has been given to how existing VLMs scale under practical GPU budgets, long-video scenarios, and compliance workflows. Challenges remain in producing reliable VLMs with growing amount of video data, reducing hallucinations and memory cost, improving quality of long video content reasoning, and addressing computational bottlenecks \citep{qu-etal-2025-look}. Industrializing VLMs for large-scale data especially using closed-source models on proprietary data is also becoming increasingly urgent.

%In this work, we centrial around the Reasrearchq uestions (RQ): namely,  VLM-based multimodal reasoning on long-form pharmaceutical videos under realistic industrial GPU constraints, we open-sourced a large-scale industrial GenAI platform architecture for natural language search for H&PS User. The platform ingested over 200,000 PDFs, 25,326 videos across 8 various formats (e.g., MP4, M4V, QuickTime, WMV, WebM, MSVideo, MPG, and 3GPP), and 888 audio files across 20+ languages. We then analyze more than 40 VLMs on two leading benchmarks (Video-MME, MMBench) and on a proprietary dataset covering 14 disease areas. Our contributions are:

This work centers on a core \textbf{Research Question (RQ)}: how to scale VLM-based multimodal reasoning on long-form pharmaceutical videos under realistic industrial GPU constraints. To address this, we open-source a large-scale GenAI platform architecture designed for natural language search for (H\&PS) users. The platform ingested over 200,000 PDFs, 25,326 videos across 8 formats (e.g., MP4, M4V, MSVideo, etc.), and 888 audio files spanning 20+ languages. We evaluate more than 40 VLMs using leading benchmarks (Video-MME, MMBench) and our proprietary dataset covering 14 disease areas. Our contributions including:

\begin{itemize}
\item A industrial multimodal architecture framework (in Figure~\ref{fig: architecture}) for scalable ingestion, captioning, and retrieval of large amount of data.

\item  Four key findings from industrial deployment: (1) multimodality boosts VLM performance across most tasks (8/12); (2) attention mechanisms show GPU-specific trade-offs; (3) both open- and closed-source models struggle with temporal alignment and keyframe reasoning; and (4) long-video splitting \& compression is more error-prone rather than efficient.

\item Extension of Video-MME with new subtasks (summarization, keyframe evaluation) along with newly designed evaluation schema using the Knowledge Graph, shown in Alg.~\ref{alg: Fruchterman-Reingold}, \ref{alg: Dijkstra}, \ref{alg: knowledgegraph}.
\end{itemize}

\section{Related Work}

Vision-Language Models (VLMs) have advanced significantly, with several state-of-the-art models demonstrating strong performance across multimodal tasks \citep{fu2024video}. Qwen-VL \citep{Qwen-VL}, improved vision–language alignment through extensive pretraining and data scaling. 
Recent multimodal Open-source models, such as CLIP \citep{radford2021learning}, BLIP \citep{li2022blip} , FLAVA \citep{singh2022flava}, and OFA \citep{wang2022ofa}, have demonstrated strong zero-shot and fine-tuned performance in various vision-language tasks across recognition, captioning, and retrieval. More recent VLMs—including Gemini Pro/Flash \citep{GeminiTeam2024Gemini15}, AdaReTaKe \citep{Lourentzou2021AdaReNet}, and Qwen2-VL \citep{Qwen-VL}, have pushed the boundaries of video-based multi-modal understanding by improving reasoning, temporal alignment, and multi-modal fusion strategies. As shown in recent leader-boards \citep{fu2024video}, Qwen2-VL \citep{Qwen-VL} achieves state-of-the-art results in long \& short video caption tasks, reinforcing its role as a leading open-source VLMs.

Meanwhile, several benchmarks have been proposed to systematically evaluate the capabilities of MLLMs. Video-MME \citep{fu2024video} introduces an evaluation framework specifically designed for VLM, evaluating their understanding of dynamic and multimodal content. Additionally, MME \citep{fu2023mme} provides a comprehensive benchmark for the evaluation of more general multimodal LLMs, while MME-Survey  \citep{fu2024mme} offers a detailed review of existing evaluation methodologies. The MME-RealWorld benchmark \citep{zhang2024mme} further extends this evaluation to real-world, high-resolution scenarios, testing the robustness and generalization of multimodality beyond synthetic datasets. 

%Despite these advances, challenges remain in scaling VLMs under GPU constraints and in handling temporal reasoning for long videos.

What's more, efficient attention mechanisms have been crucial for scaling large multimodal models \citep{HuggingFaceTransformersPerformance}. FlashAttention \citep{dao2023flashattention2} and Scaled Dot-Product Attention (SDPA) \citep{vaswani2017attention} have played significant roles in improving efficiency in transformer-based architectures. FlashAttention reduces memory overhead and computational costs by optimizing key-query-value matrix operations, making it well-suited for large-scale applications. Similarly, SDPA, widely implemented in frameworks like Hugging Face's Transformers, optimizes inference performance on GPUs, particularly with hardware accelerations e.g. AMD ROCm and NVIDIA TensorRT \citep{HuggingFaceTransformersPerformance}.

\section{Dataset And Experimental Settings}

\begin{table}[th]
    \centering
    \caption{Statistics of Our Property Dataset.}
    \resizebox{0.5\textwidth}{!}{  % Resize the table to fit the page width
    \begin{tabular}{ll}
        \toprule
        \textbf{Category} & \textbf{Details} \\
        \midrule
        VLM Models Covered   &  42, GPT series include GPT-4, Gemini 1.5 Pro. 2.0-Flash,\\ & Qwen-7B-VL, Qwen-72B, Owen VL Max, LLaVA-Video, \\ & Oryx-1.5, InternVL 2.5, Aria \\
 & VideoLLaMA series, VideoChat Flash, NViLA, GPT-4o, \\
 & Claude 3.5 Sonnet, TimeMarker, MiniCPM-V 3.2, \\
& MiniCPM-V 2.6, InternVL series, ST-LLaMA \\
& Video-XL, VITA-1.5, Kangaroo, Video-CCAM,  \\
& ShareGemini, SIMM, Chat-Uni-VL 1.5, \\
& VideoChat2 Mistral, ShareGPT-4VVideo, \\
        ALM Models Covered  & Whisper-turbo and Whisper-large V2 \\
        Number of Videos    & Over \textbf{25,326}. \\
        Number of Audios     & Over \textbf{888}. \\
        Covered Variants     & Over \textbf{14} Diseases Ares. From Nephrology, Ophthalmology, Oncology, ...\\ & to Hematology, Immunology, Dermatology. \\
        Covered video format Types     & \textbf{8}. MP4, M4V, QuickTime, WMV, WebM, MSVideo, MPG, and 3GPP \\
        Covered audio format Types     & \textbf{4}. '.mp3', '.wav', '.m4a', '.flac'  \\
        Covered Video Lengths      & < 2 mins to over 3 hours\\
        Language Types       & Over 20 languages, including German, Italian, English, Mandarin \\&Hokkien, Hindi, Korean, French, Dutch, Spanish, and more. \\   
        \bottomrule
    \end{tabular}
    } \label{tab: RICI data}
\end{table}

Here, we primarily adopt two well-established MLLM benchmarks, Video-MME \citep{fu2024video} and MMBench \citep{MMBench}, along with evaluations of more than 40+ VLMs, as well as our pharmacy property dataset shown in Table~\ref{tab: RICI data}, and Figure ~\ref{fig: BenchmarkData}. The benchmarks allow standardized comparisons, while our dataset provides a realistic testbed for long-form industrial content.

Video-MME \citep{fu2024video} is the first full-spectrum multi-modal evaluation benchmark designed specifically for video-based MLLMs. It stands out from existing benchmarks with several key features: (1) Diversity in video types, covering six primary visual domains with 30 subfields to ensure broad scenario generalizability; (2) Temporal coverage, including short-, medium-, and long-term videos ranging from 11 seconds to 1 hour. Video-MME includes 900 manually selected videos totaling 254 hours, annotated with 2,700 question-answer pairs. 

%Extensive evaluations of state-of-the-art MLLMs, including GPT-4 series \citep{instructGPT}, Gemini 1.5 Pro \citep{GeminiTeam2024Gemini15}, InternVL-Chat-V1.5 \citep{chen2024how}, and LLaVA-NeXT-Video \citep{zhang2024llavanextvideo}, reveal that Gemini 1.5 Pro is the best-performing commercial model and Qwen the best Open-source model. 

MMBench \citep{MMBench} is designed to assess our findings across diverse visual understanding tasks, including video recognition, captioning, visual question answering (VQA), and reasoning. Furthermore, our findings are tested from our industry data, includes 25,326 videos across eight formats (MP4, M4V, QuickTime, WMV, WebM, MSVideo, MPG, 3GPP) and 888 audio files across four formats (.mp3, .wav, .m4a, .flac). The content spans 14 disease areas, including oncology, hematology, immunology, ophthalmology, neuroscience, dermatology, nephrology, and respiratory disease, covering >20 languages. Video lengths range from under 2 minutes to more than 3 hours, reflecting the diversity of clinical trial recordings, medical lectures, and patient education materials. 

We benchmarked 42 VLMs / ALMs, including GPT-4 series \citep{instructGPT}, Gemini 1.5 Pro \citep{GeminiTeam2024Gemini15}, InternVL-Chat-V1.5 \citep{chen2024how}, and LLaVA-NeXT-Video \citep{zhang2024llavanextvideo}, and Whisper etc. Each model was tested under default configurations and resource constrained GPU environments (NVIDIA A100 vs. A10G). Prompts are listed in Appendix Table~\ref{tab:hypo}.

\section{Business And Technical Impact}

Finding reliable content remains a major challenge for healthcare professionals (HCPs) and patients. Traditional search methods are inefficient, leading to under-utilized assets and duplicated content creation. VLM-based NL search system streamlines discovery, reuse, and accessibility of video content. In production pilots, the system reduced the time required to create patient-facing materials by \textit{\textbf{66\%}}, accelerating workflows and improving consistency. On Video-MME, the end-to-end processing time, from voice abstraction by Ffmpeg, Whisper Turbo transcription to Bedrock LLM improvement on VLM video summary \& captions storage in the database, averaged 2.2 minutes per longer video, 1.7 minutes per medium-length video, 1.6 minutes per short video. Compared to manual inspection (252.5 hours for the dataset), this represents a \textit{\textbf{94.4\%}} reduction in effort for long videos and a \textit{\textbf{88\%}} reduction for overall video categories. Scaling to the entire dataset implies a savings of approximately c.a. \textit{\textbf{224.3 hours}}.

\section{Main Results}
%update 06th  March jonas

% ---------- Figure 3 (2×2) ----------
\begin{figure*}[t]
  \centering
  \newcommand{\figw}{0.49\textwidth}   % use \textwidth for figure* (spans both columns)
  \newcommand{\figH}{4.6cm}

  % Row 1
  \begin{subfigure}[t]{\figw}
    \centering
    \includegraphics[height=\figH,width=\linewidth,keepaspectratio]
      {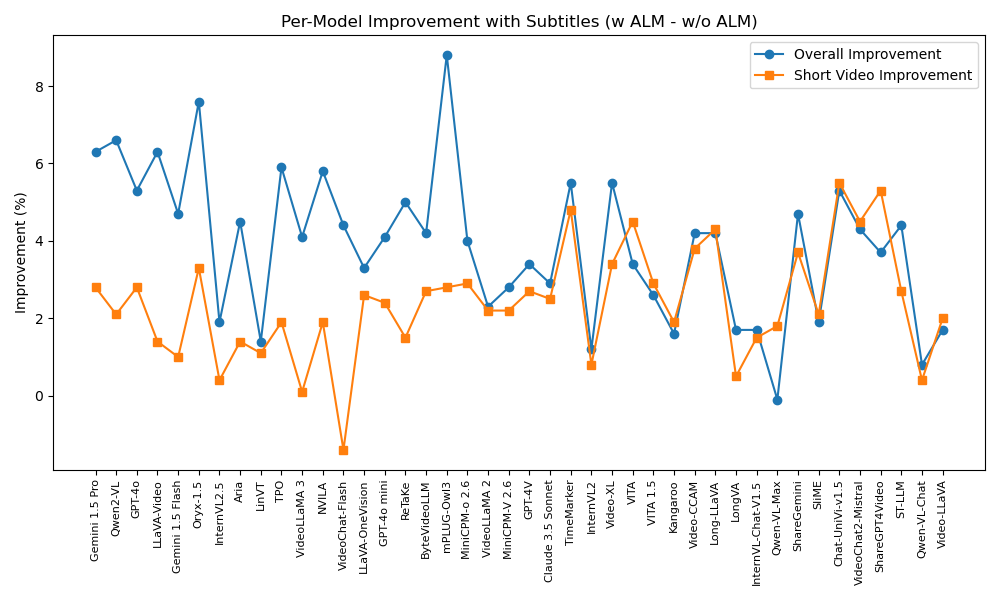}
    \subcaption{Per-model improvement with ALM (overall vs. short).}
    \label{fig:3a}
  \end{subfigure}\hfill%
  \begin{subfigure}[t]{\figw}
    \centering
    \includegraphics[height=\figH,width=\linewidth,keepaspectratio]
      {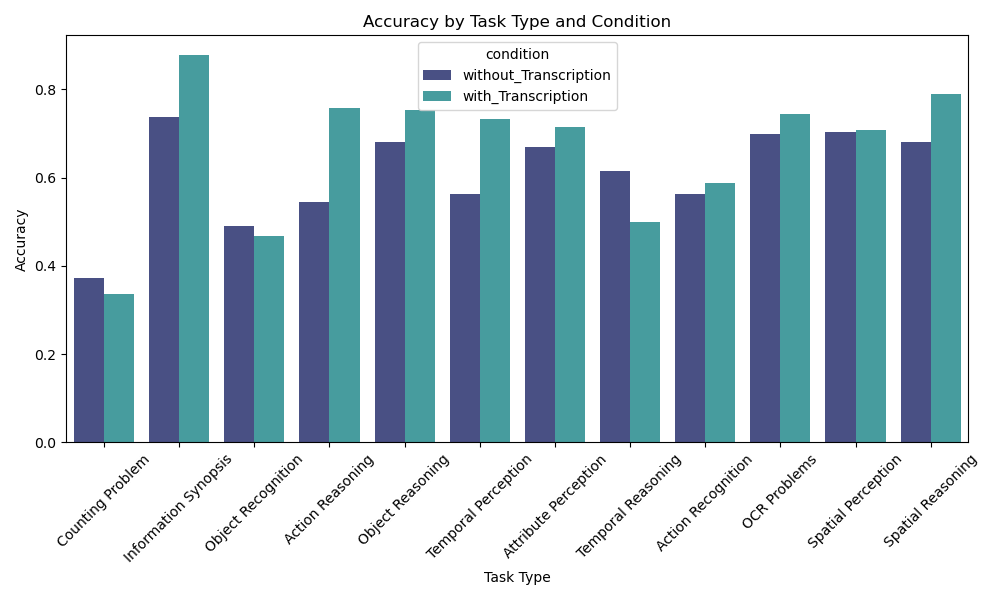}
     \subcaption{Accuracy by Sub task Type with/without transcription.}
    \label{fig:3b}
  \end{subfigure}

  \vspace{4pt}

  % Row 2
  \begin{subfigure}[t]{\figw}
    \centering
    \includegraphics[height=\figH,width=\linewidth,keepaspectratio]
      {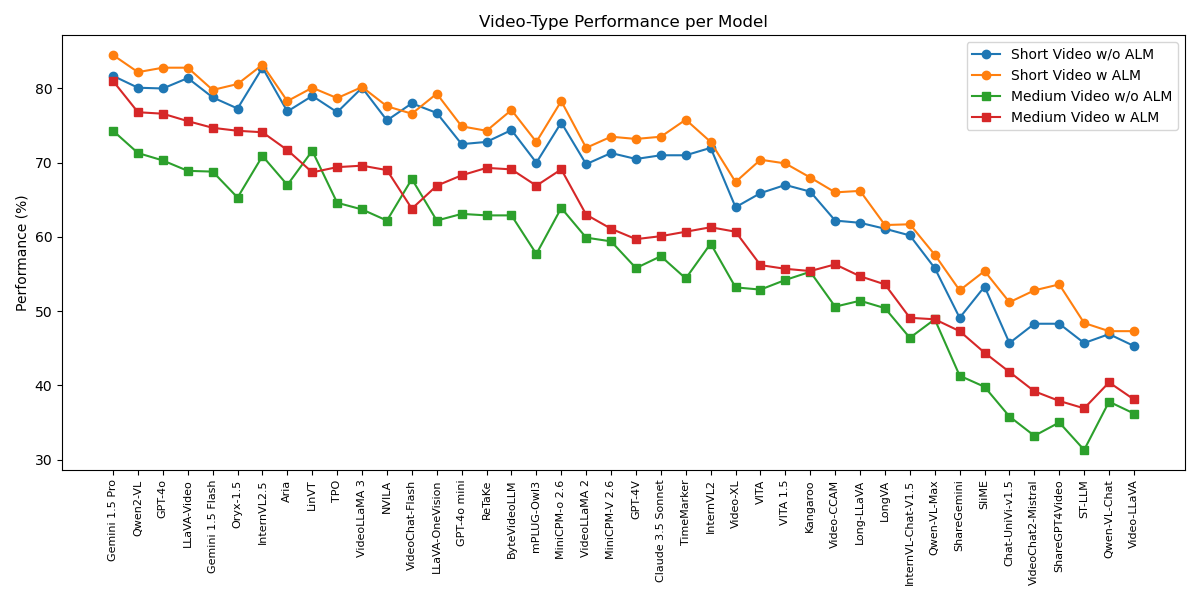}
    \subcaption{Video-type performance per model.}
    \label{fig:3c}
  \end{subfigure}\hfill%
  \begin{subfigure}[t]{\figw}
    \centering
    \includegraphics[height=\figH,width=\linewidth,keepaspectratio]
      {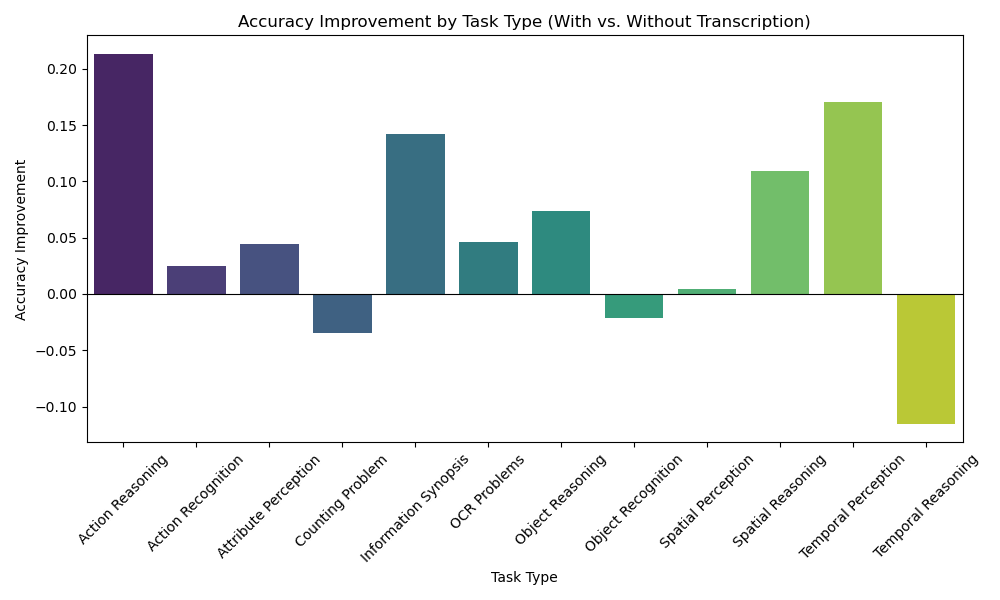}
    \subcaption{Accuracy by Sub task Type with/without transcription.}
    \label{fig:3d}
  \end{subfigure}

  \caption{\textbf{Multimodality matters.} Combining metadata and voiceovers using LLM/ALMs could improve VLM summaries and understanding. Especially effective for longer videos, and for tasks like action recognition, object reasoning, and OCR, but may negatively impact temporal reasoning tasks.}
  \label{fig:finding1}
\end{figure*}

In this section, we first evaluate with the top two benchmarks, Video-MME \citep{fu2024video} and MMBench \citep{MMBench}, as well as our proprietary dataset for multimodal vision language models (MVLMs), focusing on widely recognized 40 + vision language models (VLMs). Results are structured around four interesting findings.

\begin{table}[t]
\centering
\small
% tighten column padding a bit (scoped to this table)
\setlength{\tabcolsep}{4.5pt}
\renewcommand{\arraystretch}{1.05}

\resizebox{\linewidth}{!}{%
\begin{tabular}{lccc}
\toprule
\textbf{Task Type} & \textbf{With ALM} & \textbf{Without ALM} & $\Delta$ \\
\midrule
Action Reasoning     & 0.759 & 0.545 & +0.213 \\
Action Recognition   & 0.589 & 0.564 & +0.025 \\
Attribute Perception & 0.671 & 0.646 & +0.025 \\
Counting Problem     & 0.337 & 0.372 & --0.035 \\
Information Synopsis & 0.879 & 0.737 & +0.142 \\
OCR Problems         & 0.744 & 0.698 & +0.046 \\
Object Recognition   & 0.469 & 0.490 & --0.021 \\
Spatial Perception   & 0.708 & 0.704 & +0.005 \\
Spatial Reasoning    & 0.789 & 0.680 & +0.108 \\
Temporal Perception  & 0.733 & 0.563 & +0.171 \\
Temporal Reasoning   & 0.500 & 0.615 & --0.115 \\
\midrule
\textbf{Average}     & 0.683 & 0.626 & +0.057 \\
\bottomrule
\end{tabular}%
}

\caption{Effect of multimodality on VLM performance (Video-MME) and task-level impact of audio transcription. Adding ALM-based voice transcriptions improves 8/12 task domains. Largest gains: Action Reasoning (+0.213), Information Synopsis (+0.142), Temporal Perception (+0.171). Negative effects: Counting (--0.035), Object Recognition (--0.021), Temporal Reasoning (--0.115).}
\label{tab:multimodality}
\end{table}

\begin{table}[t]
\centering
\small
\begin{tabular}{lccc}
\toprule
\textbf{Model} & \textbf{w/o} & \textbf{w/} & $\Delta$ \\
\midrule
Gemini 2.5 Pro   & 84.7 & 85.2 & +0.5 \\
Gemini 1.5 Pro   & 75.0 & 81.3 & +6.3 \\
Qwen2-VL         & 71.2 & 77.8 & +6.6 \\
GPT-4o           & 69.0 & 77.2 & +8.2 \\
LLaVA-Video      & 76.0 & 76.9 & +0.9 \\
Gemini 1.5 Flash & 72.6 & 75.0 & +2.4 \\
Oryx-1.5         & 67.3 & 74.9 & +7.6 \\
InternVL2.5      & 67.6 & 74.0 & +6.4 \\
Aria             & 70.3 & 72.1 & +1.8 \\
LinVT            & 65.6 & 71.7 & +6.1 \\
TPO              & 66.2 & 71.5 & +5.3 \\
\midrule
\textbf{Average} & 68.4 & 72.3 & +3.9 \\
\bottomrule
\end{tabular}
\caption{Model-level impact of incorporating audio transcriptions. Overall accuracy increases from 58.4\% to 62.3\% (+3.9). Gains are largest for GPT-4o (+8.2), Oryx-1.5 (+7.6), and Qwen2-VL (+6.6). Newer models such as Gemini 2.5 Pro show smaller gains (+0.5), reflecting diminishing returns as models advance.}
\label{tab:model_audio}
\end{table}

\textbf{Multimodality Matters.} \textbf{Combining metadata and voiceovers using LLMs/ALMs enhances VLM summaries and understanding}, although ALM-only pipelines can already achieve competitive content summarization. Across 42 evaluated VLMs, including GPT-4, we observe a consistent positive trend in performance when audio transcriptions are incorporated. Notably, this improvement strongly correlates with \textbf{video length}. Overall accuracy increased from 58.4\% to 62.3\% (+3.9 points). For short videos, accuracy rose from 67.7\% to 70.0\% (+2.3 points), for medium videos from 69.3\% to 74.3\% (+5.0 points), and for long videos from 61.7\% to 69.6\% (+7.9 points). As video length increases, video Q\&A Reasoning accuracy declines significantly from over 80\% on short clips to below 50\% on long-form videos, highlighting the persistent challenge of long video understanding. Incorporating subtitles and audio information mitigates this issue, yielding improvements of up to 7.9\% for long videos, as shown in Figure~\ref{fig:finding1} (a,b,c).

When further examining multimodality with sub-task improvements, we can see that it improves 8/12 task domains, such as action reasoning, action recognition, information synopsis, OCR problems, and object reasoning, with improvements ranging from 2\% to 20\%. Here, we see the benefits of voice information in understanding procedural or sequential tasks. However, \textbf{temporal-related tasks remain challenging}, with several models showing reduced performance when processing both audio and visual inputs simultaneously (Figure~\ref{fig:finding1}d). This suggests that current VLMs still struggle with synchronizing multimodal context over extended time spans. Cross-modal attention mechanisms are typically optimized for semantic alignment rather than temporal synchronization. Our results suggest that while multimodal inputs improve overall understanding, they can introduce temporal noise, leading to higher rates of misaligned timestamps and incorrect segment boundaries in long videos (see Appendix Table~\ref{table: outputresults}). Additionally, we also noticed a decline in completeness of answered questions, dropping from 817 to 699, as CUDA out-of-memory (OOM) issues frequently occurred when reasoning with both audio \& visual information.
%Temporal reasoning in VLMs requires precise alignment between visual tokens and audio or textual representations. 

\begin{table*}[h]
    \centering
    \caption{Comparison of speed, accuracy, and completeness across leading attention mechanism on all length videos.SDPA yields higher accuracy and 4× faster runtime on commodity GPUs (A10G), while FlashAttention favors high-end A100 GPUs.}
    \renewcommand{\arraystretch}{0.8}
    \begin{tabular}{c|c|c|c}
        \toprule
        \rowcolor{gray!20} \textbf{Experiments} & \textbf{Processing Time} & \textbf{Total Answered (\%)} & \textbf{Correct Answered (\%)} \\
        \midrule
         SDPA (0.1 FPS) & 4h 37m 2s & 37\% & \textcolor{red!100}{58.73\%} \\
         SDPA (0.01 FPS) & 2h 12m 37s & 87\% & 48.40\% \\
         FlashAttention (0.1 FPS) & 7h 40m 17s & 100\% & \textcolor{red!50}{54.81\%} \\
         FlashAttention (0.01 FPS) & 1h 50m 11s & 100\% & 48.55\% \\
        \bottomrule
    \end{tabular}\label{table: finding 2}
\end{table*}

\begin{figure*}[h]
    \centering
    \begin{minipage}{1\textwidth}
        \centering
        \includegraphics[width=1\linewidth]{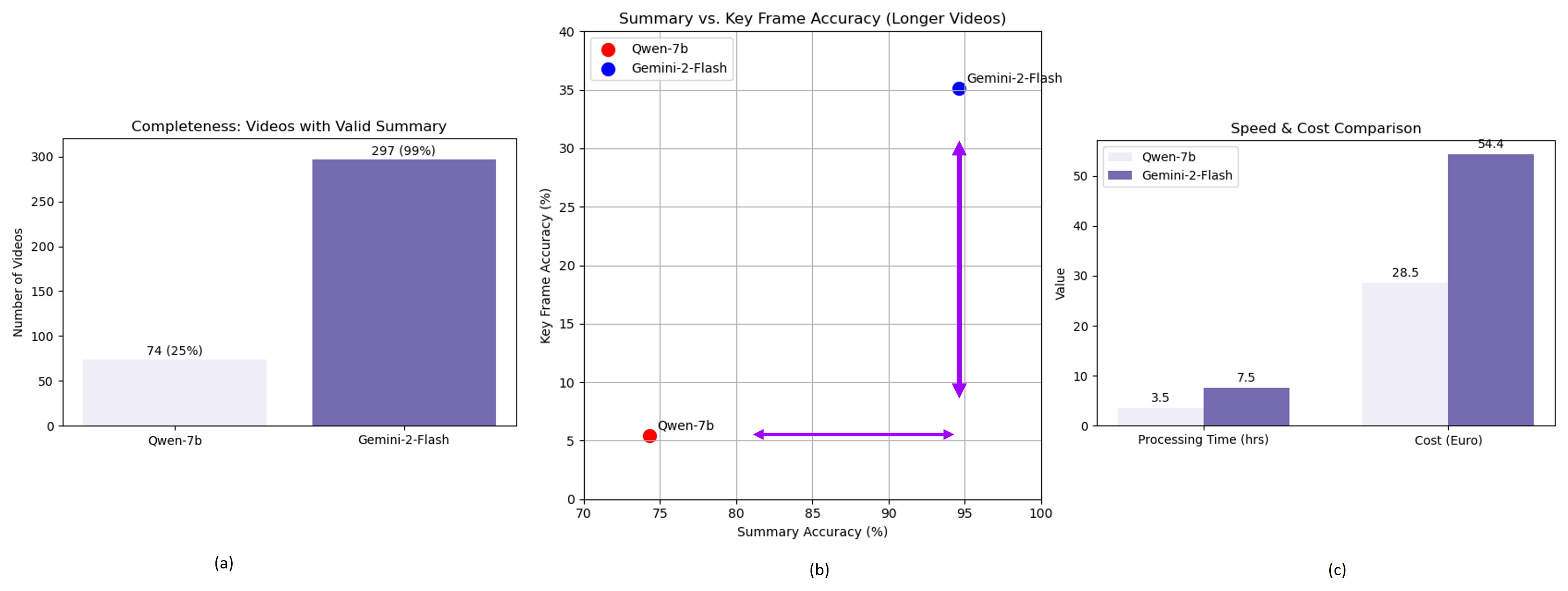}  
    \end{minipage}%
    \hfill
    \vspace{-10pt}
    \caption{Time Localization Challenges for Open-Source and Closed-Source VLMs. Both top open-source and commercial models struggle with key frame detection, showing low accuracy (5-35\%) and incorrect timestamps. Summaries are much more accurate, ranging from 75-95\%. 
 }
    \label{fig: finding 3}
\end{figure*}

\textbf{Attention Mechanisms Matter.} The VLM community has shown growing interest in FlashAttention-2~\citep{dao2023flashattention2}, as it provides an efficient implementation of the standard attention mechanism by parallelizing computation across sequence length and optimizing GPU memory usage. While FlashAttention significantly accelerates inference, most leading VLMs are trained and benchmarked on high-end GPUs such as the NVIDIA A100, which inherently benefits from optimized tensor operations and high memory bandwidth. 

However, \textbf{FlashAttention-2 also exhibits strict GPU architecture dependencies}, requiring \texttt{Torch}~$>$~3.6 and being optimized primarily for NVIDIA A100 or select AMD GPUs (e.g., Instinct MI210, MI250). In addition, dtype constraints limit its operation to \texttt{fp16} or \texttt{bf16}, which restricts deployment in many industrial environments. \textbf{Deploying these mechanisms on standardized A10G Tensor Core GPUs—rather than on large A100 UltraCluster setups—introduces major cost and compatibility challenges.} For instance, an A100 UltraCluster costs approximately \$40.96 per hour, whereas AWS G5 instances (A10G GPUs) range from \$5.67 to \$8.14 per hour, making A100-based inference roughly eight times more expensive. Empirically, FlashAttention-2 performs best on A100 GPUs and remains a popular choice for research benchmarks. In contrast, Scaled Dot-Product Attention (SDPA)~\citep{vaswani2017attention} proves to be a more practical option for A10G-based deployments, offering up to a 4$\times$ speed improvement. Notably, SDPA achieved higher accuracy (58.73\%) than FlashAttention (54.81\%) when evaluated at FPS~=~0.1 on the 2,700 Q\&A Video-MME benchmark. SDPA completed inference in 4~hours and 37~minutes, compared to 7~hours and 40~minutes for FlashAttention-2, as summarized in Table~\ref{table: finding 2}, \ref{table:fps}, \ref{table: fps2}. These findings suggest that \textbf{VLMs exhibit strong GPU architecture and FPS dependencies}, implying that the optimal attention mechanism varies with the hardware environment. SDPA ~\citep{vaswani2017attention} can also invoke FlashAttention and other memory-efficient attention kernels when needed, and native SDPAsupport is now expanding in the \texttt{Transformers} library. This underscores that selecting a well-matched--rather than the newest--attention mechanism can yield superior efficiency under realistic industrial GPU constraints.

\textbf{Time Localization Remains Challenging:} We evaluated VLM performance across three dimensions: speed, cost, and output completeness. Our findings indicate that Gemini 2.5 Flash is slower and more expensive, requiring 7--8 hours to process 300 videos at a cost of Euro 46.73 on Google Cloud. In contrast, Qwen-7B~\citep{Qwen-VL} demonstrated substantially faster processing, completing each video in 1 mins on an AWS EC2 \texttt{g5.24xlarge} instance, priced at~\$8.14~per hour. However, when running at 0.01~FPS with SDPA attention, Qwen-7B required approximately 3--4~hours to process 300~videos, as shown in Figure~~\ref{fig: finding 3}. In terms of \textbf{output completeness}, Gemini~2.5~Flash outperformed Qwen-7B, generating valid summaries for 297/300 videos (with acc.94.6\%), while Qwen-7B produced only 74 valid summaries (with acc.74.3\%). Additionally, \textbf{key frame extraction remained challenging for both models}, with Gemini~2.5~Flash achieving a keyframe accuracy of~35.1\%~(26/74), while Qwen-7B reached only~5.4\%~(4/74). For long videos, Gemini~2.5~Flash required roughly 3 mins per video, while Qwen-7B remained below one minute per video. Despite these differences, \textbf{both leading open- and closed-source VLMs struggle with accurate temporal reasoning}, particularly in maintaining keyframe alignment and coherence across long video sequences (see Table ~\ref{table: outputresults}). This highlights a persistent limitation in time-dependent video understanding for current VLM architectures. 

%Many VLMs rely on fixed or softly bounded token budgets for visual inputs, often leading to the loss of fine-grained temporal cues. 

%Our experiments show that as video length increases, keyframe localization accuracy degrades disproportionately compared to semantic understanding, suggesting that temporal ordering information is discarded earlier than high-level visual concepts.

\textbf{Trade-off of Long Video Splitting.} Processing long videos poses significant challenges, particularly on GPU-constrained instances. Out-of-memory (OOM) errors occur frequently, even when reducing frames per second (FPS) or lowering video resolution. A common mitigation strategy involves video splitting, compression, cutting. However, our analysis shows that these methods do not yield meaningful speed improvements and instead introduce additional challenges related to temporal alignment. \textbf{Splitting videos into multiple segments disrupts temporal consistency, making it increasingly difficult for VLMs to maintain coherent event sequences.} Moreover, compressing video files using standard tools such as FFmpeg requires substantial preprocessing time. Although shorter segments can be processed individually, the lack of contextual continuity causes VLMs to \textbf{focus on superficial cues}—such as logos, text color, or general stylistic attributes—rather than on semantic content. This leads to redundant descriptions and weaker keyframe alignment, as shown in Table~\ref{table: sampleoutput}. 

%We often observe that misalignment errors often occur near segment boundaries, indicating that temporal events shorter than the sampling interval are frequently missed.

\section{Conclusion}

%In this work, we introduce a novel industrial framework for large-scale VLM-based video processing and NL search. Unlike prior work focusing on new architectures, our contribution lies in benchmarking, scaling, and analyzing existing VLMs under realistic GPU and compliance constraints. Through our extensive evaluation, we highlight the following key findings: 1) Multimodality Matters: While current research predominantly focuses on static image inference or bi-modality, incorporating ALMs, voiceover transcriptions, and metadata improves video reasoning. 2) Attention Mechanisms Matter – Selecting right attention mechanism based on the GPU architecture can yield better efficiency. 3) Both Open \& Closed-source VLMs perform well in video summarization but continue to struggle with key frame detection and timestamp. 4) While splitting long videos into shorter segments appears to be an intuitive solution, cutting and compressing videos often consumes even more processing time and increases errors in key frame alignment and summary. Beyond these insights, our work extends Video-MME with new subtasks (summarization, keyframe extraction) and proposes a knowledge-graph evaluation schema. Future research should extend these insights beyond healthcare, exploring financial and manufacturing video domains under constrained hardware.

In this work, we introduce an industrial framework for large-scale VLM-based video processing and NL search. Unlike prior studies proposing new architectures, our contribution lies in benchmarking, scaling, and analyzing existing VLMs under realistic GPU and compliance constraints. Our evaluation yields four key findings: (1) \textbf{Multimodality matters}---incorporating ALMs, transcriptions, and metadata improves video reasoning beyond static or bi-modal setups; (2) \textbf{Attention mechanisms matter}---matching attention to GPU architecture improves efficiency; (3) \textbf{Open- and closed-source VLMs} perform well in video summarization but still struggle with keyframe detection and timestamping; and (4) \textbf{Splitting long videos} often increases runtime and misalignment errors rather than improving efficiency. Beyond these insights, we extend Video-MME with new subtasks (summarization, keyframe extraction) and propose a knowledge-graph evaluation schema. Future research should exploring financial and manufacturing video domains under constrained hardware.

\newpage
\clearpage

\section{Limitations}
We want to emphasize that this work is not intended to introduce a new VLM architecture, but rather to empirically characterize the capabilities, limitations, and failure modes of existing models under realistic long-video and hardware constraints.
A main focus of this study is the justification of the pharma specific benefits, e.g. how to scale VLMs usage in industry GPU constrained hardware setting for business application. We provide a baseline comparison of more than 42 VLMs using both the Video-MME benchmark and our proprietary dataset. Future work should extend this line of research to other regulated domains, such as financial services and manufacturing videos e.g. Figure~\ref{fig: business3}, to further validate the generalization of our solution blueprint.

\section{Acknowledgments}  
We sincerely thank Samik Adhikary and Puneet Srivastava for their sponsorship support from Roche. We also appreciate the insightful discussions and technical assistance provided by Janina Kummerfeldt, Philipp Müller, and Kathrin Schwan from Accenture, Jennifer McGuire's business support from Roche,  as well as Aditya Kumar's paper experiment support from Involead.
This platform, RICI, would not have been possible without their contributions. We further extend our gratitude to the backend engineering teams supported RICI development, as well as to the healthcare professionals (HCPs), testers, and Roche Lab users whose consistent feedback brought our Content Finder use cases to life and enabled continuous improvement.
This paper serves as our academic contribution, where we formalized experiments and evaluation methodologies using academic benchmarks. Through this work, we aim to share industry lessons learned and valuable large-scale GenAI experiments in the pharmaceutical domain.
\bibliography{custom}

@article{Li2021MedicalDatasets,
  title = {A Systematic Collection of Medical Image Datasets for Deep Learning},
  author = {Li, Johann and Zhu, Guangming and Hua, Cong},
  year = {2021},
  journal = {arXiv preprint arXiv:2106.12864}
}

@article{Najjar2023RadiologyReview,
  title = {Redefining Radiology: A Review of Artificial Intelligence},
  author = {Najjar, R. et al.},
  year = {2023},
  journal = {PMC}
}

@article{PintoCoelho2023AIImaging,
  title = {How Artificial Intelligence Is Shaping Medical Imaging},
  author = {Pinto-Coelho, L. et al.},
  year = {2023},
  journal = {PMC}
}

@article{Wang2025UAVSurvey,
  title = {A survey of unmanned aerial vehicles and deep learning in precision agriculture},
  author = {Wang, D. et al.},
  year = {2025},
  journal = {ScienceDirect}
}

@article{Strzepek2023AgricultureUAV,
  title = {Quantitative and Qualitative Analysis of Agricultural Fields with UAVs},
  author = {Strzępek, K. et al.},
  year = {2023},
  journal = {PMC}
}

@article{fu2024video,
  title={Video-MME: The First-Ever Comprehensive Evaluation Benchmark of Multi-modal LLMs in Video Analysis},
  author={Fu, Chaoyou and Dai, Yuhan and Luo, Yondong and Li, Lei and Ren, Shuhuai and Zhang, Renrui and Wang, Zihan and Zhou, Chenyu and Shen, Yunhang and Zhang, Mengdan and others},
  journal={arXiv preprint arXiv:2405.21075},
  year={2024}
}

@inproceedings{li-etal-2025-llms,
    title = "How {LLM}s React to Industrial Spatio-Temporal Data? Assessing Hallucination with a Novel Traffic Incident Benchmark Dataset",
    author = "Li, Qiang  and
      Tan, Mingkun  and
      Zhao, Xun  and
      Zhang, Dan  and
      Zhang, Daoan  and
      Lei, Shengzhao  and
      Chu, Anderson S.  and
      Li, Lujun  and
      Kamnoedboon, Porawit",
    editor = "Chen, Weizhu  and
      Yang, Yi  and
      Kachuee, Mohammad  and
      Fu, Xue-Yong",
    booktitle = "Proceedings of the 2025 Conference of the Nations of the Americas Chapter of the Association for Computational Linguistics: Human Language Technologies (Volume 3: Industry Track)",
    month = apr,
    year = "2025",
    publisher = "Association for Computational Linguistics",
    url = "https://aclanthology.org/2025.naacl-industry.4/",
    pages = "36--53"
   
}

@article{Saeidi2022STAR,
  title   = {Autonomous robotic laparoscopic surgery for intestinal anastomosis},
  author  = {Saeidi, H. and et al.},
  journal = {Science Robotics},
  year    = {2022},
  doi     = {10.1126/scirobotics.abj2908},
  url     = {https://www.science.org/doi/10.1126/scirobotics.abj2908}
}

@article{Borgli2020HyperKvasir,
  title   = {HyperKvasir, a comprehensive multi-class image and video dataset for gastrointestinal endoscopy},
  author  = {Borgli, Hanna and et al.},
  journal = {Scientific Data},
  year    = {2020},
  doi     = {10.1038/s41597-020-00622-y},
  url     = {https://www.nature.com/articles/s41597-020-00622-y}
}

@misc{FDADEN180001,
  title        = {IDx-DR: De Novo Summary (DEN180001)},
  howpublished = {\url{https://www.accessdata.fda.gov/cdrh_docs/reviews/DEN180001.pdf}},
  year         = {2018},
  note         = {U.S. Food and Drug Administration De Novo classification for autonomous DR detection}
}

@review{Oshika2024AIOph,
  title   = {Artificial Intelligence Applications in Ophthalmology},
  author  = {Oshika, T. and et~al.},
  journal = {Japanese Journal of Ophthalmology (review; open-access PMC)},
  year    = {2024},
  url     = {https://pmc.ncbi.nlm.nih.gov/articles/PMC11799668/}
}

@review{Xu2024DRReview,
  title   = {The application of artificial intelligence in diabetic retinopathy: progress and challenges},
  author  = {Xu, X. and et~al.},
  journal = {Ophthalmology Science (review; open-access PMC)},
  year    = {2024},
  url     = {https://pmc.ncbi.nlm.nih.gov/articles/PMC11543434/}
}

@misc{ISICArchive,
  title        = {ISIC: International Skin Imaging Collaboration Archive},
  howpublished = {\url{https://www.isic-archive.com/}},
  note         = {Dermoscopic/clinical \emph{image} repository widely used for AI benchmarking}
}

@article{Yilmaz2024DERM12345,
  title   = {DERM12345: A Large, Multisource Dermatoscopic Skin Lesion Dataset},
  author  = {Yilmaz, A. and et~al.},
  journal = {Scientific Data},
  year    = {2024},
  doi     = {10.1038/s41597-024-04104-3},
  url     = {https://www.nature.com/articles/s41597-024-04104-3}
}

@review{Behara2024DermAI,
  title   = {AI in dermatology: a comprehensive review into skin disease identification},
  author  = {Behara, K. and et~al.},
  journal = {Frontiers in Medicine},
  year    = {2024},
  url     = {https://pmc.ncbi.nlm.nih.gov/articles/PMC11784784/}
}

@article{Mahmoud2024CatStep,
  title   = {CatStep: Automated Cataract Surgical Phase Classification Using Deep Learning},
  author  = {Mahmoud, O. and et~al.},
  journal = {Ophthalmology Science},
  year    = {2024},
  url     = {https://www.ophthalmologyscience.org/article/S2666-9145%2823%2900137-9/pdf}
}

@article{Ghamsarian2024Cataract1K,
  title   = {Cataract-1K: A large-scale cataract surgery video dataset for surgical workflow analysis},
  author  = {Ghamsarian, N. and et~al.},
  journal = {Scientific Data},
  year    = {2024},
  doi     = {10.1038/s41597-024-03193-4},
  url     = {https://www.nature.com/articles/s41597-024-03193-4}
}

@article{Fatima2024DigPathReview,
  title   = {A Comprehensive Review of Advances in Digital Pathology},
  author  = {Fatima, Ghazala and Khan, Imran and Farooq, Saeed and others},
  journal = {Diagnostics},
  year    = {2024},
  url     = {https://pmc.ncbi.nlm.nih.gov/articles/PMC11573928/}
}

@article{TanGarcia2025CompPath,
  title   = {Computational Pathology in the Age of Artificial Intelligence},
  author  = {Tan-Garcia, Amanda and others},
  journal = {Histopathology},
  year    = {2025},
  url     = {https://pmc.ncbi.nlm.nih.gov/articles/PMC12451823/}
}

@article{Huang2023npjDM,
  title   = {Challenges and prospects of visual contactless physiological monitoring},
  author  = {Huang, Bin and others},
  journal = {npj Digital Medicine},
  year    = {2023},
  url     = {https://www.nature.com/articles/s41746-023-00973-x}
}

@inproceedings{Du2018UAVDT,
  title     = {The Unmanned Aerial Vehicle Benchmark: Object Detection and Tracking},
  author    = {Du, Dawei and Qi, Yuankai and Yu, Hongyang and Yang, Yifan and Duan, Kaiwen and Li, Guorong and Zhang, Weigang and Huang, Qingming and Tian, Qi},
  booktitle = {Proceedings of the European Conference on Computer Vision (ECCV)},
  year      = {2018},
  pages     = {370--386},
  url       = {https://openaccess.thecvf.com/content_ECCV_2018/html/Dawei_Du_The_Unmanned_Aerial_ECCV_2018_paper.html}
}

@inproceedings{Zhang2022ByteTrack,
  title     = {ByteTrack: Multi-Object Tracking by Associating Every Detection Box},
  author    = {Zhang, Yifu and Sun, Peize and Jiang, Yi and Yu, Dongdong and Weng, Fucheng and Yuan, Zehuan and Luo, Ping and Liu, Wenyu and Wang, Xinggang},
  booktitle = {Computer Vision -- ECCV 2022},
  series    = {Lecture Notes in Computer Science},
  volume    = {13682},
  pages     = {1--21},
  publisher = {Springer, Cham},
  year      = {2022},
  doi       = {10.1007/978-3-031-20047-2_1},
  url       = {https://link.springer.com/chapter/10.1007/978-3-031-20047-2_1}
}

@article{Wei2023GlobalPM25,
  title   = {First close insight into global daily gapless 1 km PM\textsubscript{2.5} pollution, variability, and health impact},
  author  = {Wei, Jing and Li, Zhanqing and Lyapustin, Alexei and Wang, Jun and Dubovik, Oleg and Schwartz, Joel and Sun, Lin and Li, Chi and Liu, Song and Zhu, Tong and others},
  journal = {Nature Communications},
  year    = {2023},
  volume  = {14},
  pages   = {8349},
  doi     = {10.1038/s41467-023-43862-3},
  url     = {https://www.nature.com/articles/s41467-023-43862-3}
}

@misc{FDA_AIML_List,
  title        = {Artificial Intelligence (AI/ML)-Enabled Medical Devices},
  howpublished = {\url{https://www.fda.gov/medical-devices/software-medical-device-samd/artificial-intelligence-enabled-medical-devices}},
  note         = {Official FDA list of AI-enabled medical devices; Accessed 2025-10-07}
}

@misc{ACR_AICentral,
  title        = {AI Central (American College of Radiology) -- FDA-cleared imaging AI directory},
  howpublished = {\url{https://aicentral.acrdsi.org/}},
  note         = {Over 200 FDA-cleared imaging AI products cataloged; Accessed 2025-10-07}
}

@article{Haque2025rPPGBias,
  title   = {Demographic bias in public remote photoplethysmography datasets},
  author  = {Haque, Md. and others},
  journal = {npj Digital Medicine},
  year    = {2025},
  url     = {https://pmc.ncbi.nlm.nih.gov/articles/PMC12491395/}
}

@review{BJSOpen2024SurgVideo,
  title   = {Automated analysis of operative video in surgical training: a scoping review},
  author  = {Dick, L. and et~al.},
  journal = {BJS Open},
  year    = {2024},
  url     = {https://academic.oup.com/bjsopen/article/8/5/zrae124/7823786}
}

@article{Ali2023PolypGen,
  title   = {A multi-centre polyp detection and segmentation dataset for generalisability assessment (PolypGen)},
  author  = {Ali, Sharib and Jha, Debesh and Ghatwary, Noha and et al.},
  journal = {Scientific Data},
  year    = {2023},
  doi     = {10.1038/s41597-023-01981-y},
  url     = {https://www.nature.com/articles/s41597-023-01981-y}
}

@article{Rios2023Cholec80CVS,
  title   = {Cholec80-CVS: An open dataset with an evaluation of Strasberg’s Critical View of Safety for AI},
  author  = {R{\'\i}os, Manuel Sebasti{\'a}n and et al.},
  journal = {Scientific Data},
  year    = {2023},
  doi     = {10.1038/s41597-023-02073-7},
  url     = {https://www.nature.com/articles/s41597-023-02073-7}
}

@inproceedings{yanuka-etal-2025-bridging,
    title = "Bridging the Visual Gap: Fine-Tuning Multimodal Models with Knowledge-Adapted Captions",
    author = "Yanuka, Moran  and
      Ben-Kish, Assaf  and
      Bitton, Yonatan  and
      Szpektor, Idan  and
      Giryes, Raja",
    editor = "Chiruzzo, Luis  and
      Ritter, Alan  and
      Wang, Lu",
    booktitle = "Proceedings of the 2025 Conference of the Nations of the Americas Chapter of the Association for Computational Linguistics: Human Language Technologies (Volume 1: Long Papers)",
    month = apr,
    year = "2025",
    publisher = "Association for Computational Linguistics",
    url = "https://aclanthology.org/2025.naacl-long.527/",
    pages = "10497--10518"
    
}

@article{Qwen-VL,
  title={Qwen-VL: A Versatile Vision-Language Model for Understanding, Localization, Text Reading, and Beyond},
  author={Bai, Jinze and Bai, Shuai and Yang, Shusheng and Wang, Shijie and Tan, Sinan and Wang, Peng and Lin, Junyang and Zhou, Chang and Zhou, Jingren},
  journal={arXiv preprint arXiv:2308.12966},
  year={2023}
}

@inproceedings{radford2021learning,
  title     = {Learning Transferable Visual Models from Natural Language Supervision},
  author    = {Radford, Alec and Kim, Jong Wook and Hallacy, Chris and Ramesh, Aditya and Goh, Gabriel and Agarwal, Sandhini and Sastry, Girish and Askell, Amanda and Mishkin, Pamela and Clark, Jack and Krueger, Gretchen and Sutskever, Ilya},
  booktitle = {Proceedings of the 38th International Conference on Machine Learning (ICML)},
  pages     = {8748--8763},
  year      = {2021},
  url       = {https://proceedings.mlr.press/v139/radford21a.html}
}

@inproceedings{li2022blip,
  title     = {BLIP: Bootstrapped Language-Image Pre-training for Unified Vision-Language Understanding and Generation},
  author    = {Li, Junnan and Selvaraju, Ramprasaath R. and Gotmare, Akhilesh D. and Joty, Shafiq and Xiong, Caiming and Hoi, Steven C.H.},
  booktitle = {Proceedings of the 39th International Conference on Machine Learning (ICML)},
  pages     = {12888--12900},
  year      = {2022},
  url       = {https://proceedings.mlr.press/v162/li22n.html}
}

@inproceedings{singh2022flava,
  title     = {FLAVA: A Foundational Vision and Language Model for Generalized Multimodal Learning},
  author    = {Singh, Amanpreet and Hu, Ronghang and Goswami, Vedanuj and Mahajan, Dhruv and Wu, Xinlei and Feichtenhofer, Christoph and Darrell, Trevor and Girshick, Ross},
  booktitle = {Proceedings of the IEEE/CVF Conference on Computer Vision and Pattern Recognition (CVPR)},
  pages     = {15638--15650},
  year      = {2022},
  url       = {https://openaccess.thecvf.com/content/CVPR2022/html/Singh_FLAVA_A_Foundational_Vision_and_Language_Model_for_Generalized_Multimodal_Learning_CVPR_2022_paper.html}
}

@article{wang2022ofa,
  title     = {OFA: Unifying Architectures, Tasks, and Modalities Through a Simple Sequence-to-Sequence Learning Framework},
  author    = {Wang, Peng and Yang, An and Men, Rui and Zhou, Ming and Zhang, Yinan and Lin, Junyang and Sun, Xu and Li, Shuai and Wang, Houqiang and Si, Luo},
  journal   = {arXiv preprint arXiv:2202.03052},
  year      = {2022},
  url       = {https://arxiv.org/abs/2202.03052}
}

@article{GeminiTeam2024Gemini15,
  author    = {{Gemini Team, Google: Petko Georgiev and 1135 other authors}},
  title     = {Gemini 1.5: Unlocking multimodal understanding across millions of tokens of context},
  journal   = {arXiv preprint arXiv:2403.05530},
  year      = {2024},
  url       = {https://arxiv.org/abs/2403.05530}
}

@inproceedings{Lourentzou2021AdaReNet,
  author    = {Ismini Lourentzou and Chen Chen and ChengXiang Zhai},
  title     = {AdaReNet: Adaptive Reweighted Semi-supervised Active Learning},
  booktitle = {Proceedings of the 14th ACM International Conference on PErvasive Technologies Related to Assistive Environments (PETRA)},
  pages     = {1--8},
  year      = {2021},
  doi       = {10.1145/3453892.3461321},
  url       = {https://dl.acm.org/doi/10.1145/3453892.3461321}
}

@article{fu2023mme,
  title={MME: A Comprehensive Evaluation Benchmark for Multimodal Large Language Models},
  author={Fu, Chaoyou and Chen, Peixian and Shen, Yunhang and Qin, Yulei and Zhang, Mengdan and Lin, Xu and Yang, Jinrui and Zheng, Xiawu and Li, Ke and Sun, Xing and others},
  journal={arXiv preprint arXiv:2306.13394},
  year={2023}
}

@article{fu2024mme,
  title={MME-Survey: A Comprehensive Survey on Evaluation of Multimodal LLMs},
  author={Fu, Chaoyou and Zhang, Yi-Fan and Yin, Shukang and Li, Bo and Fang, Xinyu and Zhao, Sirui and Duan, Haodong and Sun, Xing and Liu, Ziwei and Wang, Liang and others},
  journal={arXiv preprint arXiv:2411.15296},
  year={2024}
}

@article{zhang2024mme,
  title={MME-RealWorld: Could Your Multimodal LLM Challenge High-Resolution Real-World Scenarios that are Difficult for Humans?},
  author={Zhang, Yi-Fan and Zhang, Huanyu and Tian, Haochen and Fu, Chaoyou and Zhang, Shuangqing and Wu, Junfei and Li, Feng and Wang, Kun and Wen, Qingsong and Zhang, Zhang and others},
  journal={arXiv preprint arXiv:2408.13257},
  year={2024}
}

@misc{HuggingFaceTransformersPerformance,
  author    = {Hugging Face},
  title     = {Transformers Performance Inference on GPU},
  year      = {2024},
  url       = {https://huggingface.co/docs/transformers/perf_infer_gpu_one?install=AMD},
  note      = {Accessed: 2024-03-10}
}

@article{dao2023flashattention2,
  title     = {FlashAttention-2: Faster Attention with Better Parallelism and Work Partitioning},
  author    = {Tri Dao},
  journal   = {arXiv preprint arXiv:2307.08691},
  year      = {2023},
  url       = {https://arxiv.org/abs/2307.08691}
}

@inproceedings{vaswani2017attention,
  title     = {Attention Is All You Need},
  author    = {Ashish Vaswani and Noam Shazeer and Niki Parmar and Jakob Uszkoreit and Llion Jones and Aidan N. Gomez and Lukasz Kaiser and Illia Polosukhin},
  booktitle = {Advances in Neural Information Processing Systems},
  year      = {2017},
  url       = {https://arxiv.org/abs/1706.03762}
}

@misc{openai2023chatgptintel,
  author = {{OpenAI}},
  title = {ChatGPT for Intelligence Briefing: A Case Study},
  year = {2023},
  url = {https://openai.com/research/chatgpt-intelligence-briefing},
  note = {Accessed: March 2025}
}

@misc{openai2023chatgpt,
  author = {{OpenAI}},
  title = {ChatGPT: Optimizing Language Models for Dialogue},
  year = {2024},
  url = {https://openai.com/index/chatgpt/},
  note = {Accessed: March 2025}
}

@misc{lewis2021retrievalaugmentedgenerationknowledgeintensivenlp,
      title={Retrieval-Augmented Generation for Knowledge-Intensive NLP Tasks}, 
      author={Patrick Lewis and Ethan Perez and Aleksandra Piktus and Fabio Petroni and Vladimir Karpukhin and Naman Goyal and Heinrich Küttler and Mike Lewis and Wen-tau Yih and Tim Rocktäschel and Sebastian Riedel and Douwe Kiela},
      year={2021},
      eprint={2005.11401},
      archivePrefix={arXiv},
      primaryClass={cs.CL},
      url={https://arxiv.org/abs/2005.11401}, 
}

@misc{li2025benchmarkevaluationsapplicationschallenges,
      title={Benchmark Evaluations, Applications, and Challenges of Large Vision Language Models: A Survey}, 
      author={Zongxia Li and Xiyang Wu and Hongyang Du and Huy Nghiem and Guangyao Shi},
      year={2025},
      eprint={2501.02189},
      archivePrefix={arXiv},
      primaryClass={cs.CV},
      url={https://arxiv.org/abs/2501.02189}, 
}

@misc{yang2024modifyingaienhancingessays,
      title={Modifying AI, Enhancing Essays: How Active Engagement with Generative AI Boosts Writing Quality}, 
      author={Kaixun Yang and Mladen Raković and Zhiping Liang and Lixiang Yan and Zijie Zeng and Yizhou Fan and Dragan Gašević and Guanliang Chen},
      year={2024},
      eprint={2412.07200},
      archivePrefix={arXiv},
      primaryClass={cs.HC},
      url={https://arxiv.org/abs/2412.07200}, 
}

@inproceedings{qu-etal-2025-look,
    title = "Look, Compare, Decide: Alleviating Hallucination in Large Vision-Language Models via Multi-View Multi-Path Reasoning",
    author = "Qu, Xiaoye  and
      Sun, Jiashuo  and
      Wei, Wei  and
      Liu, Daizong  and
      Dong, Jianfeng  and
      Cheng, Yu",
    booktitle = "Proceedings of the 31st International Conference on Computational Linguistics",
    month = jan,
    year = "2025",
    publisher = "Association for Computational Linguistics",
    url = "https://aclanthology.org/2025.coling-main.299/",
    
}

@inproceedings{instructGPT,
  author       = {Long Ouyang and Jeffrey Wu and Xu Jiang and Diogo Almeida and Carroll Wainwright and Pamela Mishkin and Chong Zhang and Sandhini Agarwal and Katarina Slama and Alex Ray and others},
  title        = {Training Language Models to Follow Instructions with Human Feedback},
  booktitle    = {Advances in Neural Information Processing Systems},
  volume       = {35},
  pages        = {27730--27744},
  year         = {2022}
}

@article{KnowledgeLLM,
  author  = {Karan Singhal and Shekoofeh Azizi and Tao Tu and S Sara Mahdavi and Jason Wei and Hyung Won Chung and Nathan Scales and Ajay Tanwani and Heather Cole-Lewis and Stephen Pfohl and others},
  title   = {Large language models encode clinical knowledge},
  journal = {arXiv preprint arXiv:2212.13138},
  year    = {2022}
}

@article{bard,
  author    = {Romal Thoppilan and Daniel De Freitas and Jamie Hall and Noam Shazeer and Apoorv Kulshreshtha and Heng-Tze Cheng and Alicia Jin and Taylor Bos and Leslie Baker and Yu Du and others},
  title     = {Lamda: Language Models for Dialog Applications},
  journal   = {arXiv preprint arXiv:2201.08239},
  year      = {2022}
}

@misc{zhang2024llavanextvideo,
  title     = {LLaVA-NeXT: A Strong Zero-shot Video Understanding Model},
  author    = {Zhang, Yuanhan and Li, Bo and Liu, Haotian and Lee, Yong Jae and Gui, Liangke and Fu, Di and Feng, Jiashi and Liu, Ziwei and Li, Chunyuan},
  year      = {2024},
  month     = {April},
  url       = {https://llava-vl.github.io/blog/2024-04-30-llava-next-video/}
}

@article{1,
  title={Active retrieval augmented generation},
  author={Jiang, Zhengbao and others},
  journal={arXiv preprint arXiv:2305.06983},
  year={2023}
}

@article{2,
  title={Dense passage retrieval for open-domain question answering},
  author={Karpukhin, Vladimir and others},
  journal={arXiv preprint arXiv:2004.04906},
  year={2020}
}

@article{4,
  title={Improving the domain adaptation of retrieval augmented generation (RAG) models for open domain question answering},
  author={Siriwardhana, Shamane and others},
  journal={Transactions of the Association for Computational Linguistics},
  volume={11},
  pages={1--17},
  year={2023}
}

@article{MMBench,
    author = {Liu, Yuan and Duan, Haodong and Zhang, Yuanhan and Li, Bo and
          Zhang, Songyang and Zhao, Wangbo and Yuan, Yike and Wang, Jiaqi and
          He, Conghui and Liu, Ziwei and Chen, Kai and Lin, Dahua},
    journal = {arXiv:2307.06281},
    title   = {MMBench: Is Your Multi-modal Model an All-around Player?},
    year    = {2023},
}

@article{chen2024how,
  title     = {How Far Are We to GPT-4V? Closing the Gap to Commercial Multimodal Models with Open-Source Suites},
  author    = {Chen, Zhe and Wang, Weiyun and Tian, Hao and Ye, Shenglong and Gao, Zhangwei and Cui, Erfei and Tong, Wenwen and Hu, Kongzhi and Luo, Jiapeng and Ma, Zheng and Ma, Ji and Wang, Jiaqi and Dong, Xiaoyi and Yan, Hang and Guo, Hewei and He, Conghui and Shi, Botian and Jin, Zhenjiang and Xu, Chao and Wang, Bin and Wei, Xingjian and Li, Wei and Zhang, Wenjian and Zhang, Bo and Cai, Pinlong and Wen, Licheng and Yan, Xiangchao and Dou, Min and Lu, Lewei and Zhu, Xizhou and Lu, Tong and Lin, Dahua and Qiao, Yu and Dai, Jifeng and Wang, Wenhai},
  journal   = {arXiv preprint arXiv:2404.16821},
  year      = {2024},
  url       = {https://arxiv.org/abs/2404.16821}
}

\appendix
\clearpage
\newpage
\section{Appendix}
\label{sec:appendix}

In this section we provide the supplementary compiled together with the main paper includes:
\begin{itemize}
%\item Our dataset distribution, metrics explanation used in Table~\ref{tab:9model_comparison} and Table~\ref{tab:CombinedTable};
\item Evaluation Metrics and Knowledge Graph Evaluation Schema on Algorithm~ \ref{alg: Fruchterman-Reingold}, \ref{alg: Dijkstra}, \ref{alg: knowledgegraph}, and Figure ~\ref{fig: knowledgenode};
\item Ablation study on Frame Per Second (FPS) in Table~\ref{table:fps}, Table~\ref {table: fps2};
\item Property Dataset distribution on Table~\ref{table:roche-data}, and VideoMME raw rata example on Table~\ref{table: videomme rawdata};
\item Deliverable attributes of each VLM / ALM and Metadata on Figure ~\ref{fig: deliverable};
\item The training details and hyper-parameters of experiments including prompts lists in Table~\ref{tab:hypo}, output example on Table~\ref{table: timeacc}, \ref{table: sampleoutput};
\item The business value case and area of impact of GenAI-driven Video Processing on Table~\ref{tab:business1}, Table~\ref{tab:business2} and Figure~\ref{fig: business3}.

\end{itemize}

\subsection{Evaluation Metrics}

\textbf{Assigned accuracy scores strategies in Finding 3, Time Localization Challenges in Open-Source
and Closed-Source VLMs.}

\begin{equation}
\text{Scores}_{a,g} = \frac{1}{n_a} \sum_{i=1}^{n_a} S(x)
\end{equation}
where
\[
S(x) =
\begin{cases} 
1 & \text{if } S_{a,i} = g_{a,i} \\\\
0 & \text{if } S_{a,i} != g_{a,i}
\end{cases}
\]
where $S$ is the Matching Node score, $a \in A$ refers to an Key Frame or Summary scenarios, $g$ refers to ground truth of timestamp, and $n_a$ is the total number of valid video output (e.g., if 74 videos have valid JSON outputs, we match key frames to verify timestamp accuracy and compare summary accordingly).

\subsection{Summary \& Key Frame evaluation using Knowledge Graph}

To compare the quality of video summaries generated by VLMs, we then employ a knowledge graph-based method. This is particularly useful in scenarios where human-annotated ground truth is incomplete or unavailable, such as with large-scale video datasets in industry setting.

\subsubsection{Knowledge Graph Construction}
We use the \texttt{NetworkX} library with \texttt{DiGraph} to construct the knowledge graph, \texttt{NetworkX} library encapsulated so well where:
\begin{itemize}
    \item Nodes represent extracted keyframes and conceptual entities (nouns or keywords) from the generated summaries.
    \item Edges represent semantic or temporal relationships between these concepts.
    \item The graph layout is generated using the \texttt{spring\_layout} function, which implements the Fruchterman-Reingold force-directed algorithm, as shown in Algorithm ~\ref{alg: Fruchterman-Reingold}.
\end{itemize}

\subsubsection{Mathematical Basis Behind}
The force-directed layout models the graph using physical analogies:
\begin{itemize}
    \item \textbf{Repulsion}: All nodes repel each other according to \textbf{Coulomb's law}.
    \item \textbf{Attraction}: Connected nodes attract each other like springs \textbf{(Hooke’s law)}.
\end{itemize}
These forces iteratively adjust node positions until a stable configuration is reached, visually revealing clustering and coverage. Thereby:
\begin{itemize}
    \item \textbf{Node Count}: Indicates the richness or breadth of extracted information.
    \item \textbf{Node to Node Distance}: Measures how widely concepts or key words are spread in the graph.
    \item \textbf{Distance to Central Node}: We compute shortest path lengths using Dijkstra’s algorithm to measure how far keyframe nodes are from the central summary node.as shown in Algorithm ~\ref{alg: Dijkstra}.
\end{itemize}

This is an emerging area with ongoing efforts to define metrics for summary evaluation without ground truth. Recent work from researchers at Google and Apple \cite{yanuka-etal-2025-bridging} (e.g., \textbf{\emph{Descriptiveness Recall, Contradiction Precision}}, Cosine Similarity) highlights the need for new metrics when ground truth of video summary, Key frame captions are missing.

In summary, we introduce new task domains based on open-source Video-MME \citep{fu2024video}  tasks, which previously lacked summary and keyframe subtasks due to the manual effort required for key frame localization labeling. To address this, we propose a knowledge graph approach to compare the output quality of various VLMs, as shown in Algorithm ~\ref{alg: knowledgegraph}. This approach provides visually interpretable and computationally supported method to assess summary \& key frame quality using graph-based representations, forming the basis for future work on automated evaluation metrics in GenAI applications.

\begin{algorithm}[H]
\caption{Force-Directed Graph Layout (\textbf{\emph{Fruchterman-Reingold)}} conceptually}
\begin{algorithmic}[1]
\State \textbf{Let:}
\Statex \quad $d(u, v)$ \quad $\gets$ distance between nodes $u$, $v$
\Statex \quad $k \gets C \cdot \sqrt{\frac{A}{n}}$ \Comment{$C$ is constant, $A$ is layout area, $n$ is number of nodes, K is optimal distance between nodes.}

\State \textbf{Forces:}
\Statex \quad \textbf{1. Attractive force (between connected nodes):}
\[
F_{\text{attr}}(d) = \frac{d^2}{k}
\]
\Statex \quad \textbf{2. Repulsive force (between all nodes):}
\[
F_{\text{rep}}(d) = \frac{k^2}{d}
\]
\State \textbf{Loop:} Apply forces iteratively until convergence or maximum iterations reached.
\end{algorithmic}
\label{alg: Fruchterman-Reingold}
\end{algorithm}

\begin{algorithm}[H]
\caption{\textbf{\emph{Dijkstra’s Algorithm}} for Shortest Paths}
\begin{algorithmic}[1]
\State \textbf{Input:} Directed graph $G = (V, E)$ with non-negative weights $w(u, v) \geq 0$
\State \textbf{Input:} Source node $s$
\State \textbf{Initialization:}
\Statex \quad $\text{distance}[v] \gets \infty$ for all $v \in V$
\Statex \quad $\text{distance}[s] \gets 0$
\Statex \quad Initialize a priority queue $Q$

\While{$Q$ is not empty}
    \State Extract node $u$ with minimum $\text{distance}[u]$
    \For{each neighbor $v$ of $u$}
        \State Update:
        \[
        \text{dist}[v] \gets 
        \min(\text{dist}[v], \text{dist}[u] + w(u, v))
        \]
    \EndFor
\EndWhile
\State \textbf{Output:} Shortest distances from $s$ to all $v \in V$
\end{algorithmic}
\label{alg: Dijkstra}
\end{algorithm}

\begin{algorithm}
\caption{Knowledge Graph Construction for \textbf{\emph{Summary and Key Frame Evaluation}}}
\begin{algorithmic}[1]

\State \textbf{Input:} JSON data with key frames from Gemini-2 Flash and Qwen-7B
\State \textbf{Output:} Visualized Knowledge Graph

\State \textbf{Step 1: Initialize Graph}
\State Create directed graph $G \gets$ nx.DiGraph().
Nodes (V) → Individual entities in the graph (e.g., "Gemini-2 Flash", "Snow White in rags"). Edges (E) → Directed connections between nodes (e.g., "Gemini-2 Flash" → "Key Frames").
Attributes → Additional properties of nodes/edges (e.g., color, size).
\State DiGraph G = (V, E) is defined as: V={nodes}, E={(source, target)}, where each edge has a direction.
\State \textbf{Step 2: Add Core Nodes}
\State Add node $KeyFrames$ with attributes (color: gray, size: 800)
\State Add node $VideoSummary$ with attributes (color: gray, size: 600)
%\State Add node $GeminiFlash$ with attributes (color: blue, size: 500)
%\State Add node $Qwen7B$ with attributes (color: red, size: 500)

\State \textbf{Step 3: Connect Models to Core Nodes}
\State Add edge $(e.g., GeminiFlash, KeyFrames)$
%\State Add edge $(Qwen7B, KeyFrames)$
\State Add edge $(e.g., GeminiFlash, VideoSummary)$
%\State Add edge $(Qwen7B, VideoSummary)$

\State \textbf{Step 4: Add Key Frames for Each Model}
\For{each $(timestamp, description)$ in \textbf{r.g. Gemini-2 Flash key frames}}
    \State Add node $description$ with attributes (color: light blue, size: 400)
    \State Add edge $(KeyFrames, description)$
\EndFor

\State \textbf{Step 5: Visualize the Graph}
\State Compute node layout $pos \gets$ spring\_layout($G$, seed=42)
\State Extract node colors and sizes for Knowledge Graph: Key Frames Evaluation.
%\State Draw graph with node attributes:
%   \Indent
%        \State Set node, edge colors
%        \State Set node, font sizes
%    \EndIndent
%\State Show graph with title: \textit{"Knowledge Graph: Key Frames Evaluation"}

\end{algorithmic}
\label{alg: knowledgegraph}
\end{algorithm}

\begin{figure*}[ht]
    \centering
    \begin{minipage}{1\textwidth}
        \centering
        \includegraphics[width=1\linewidth]{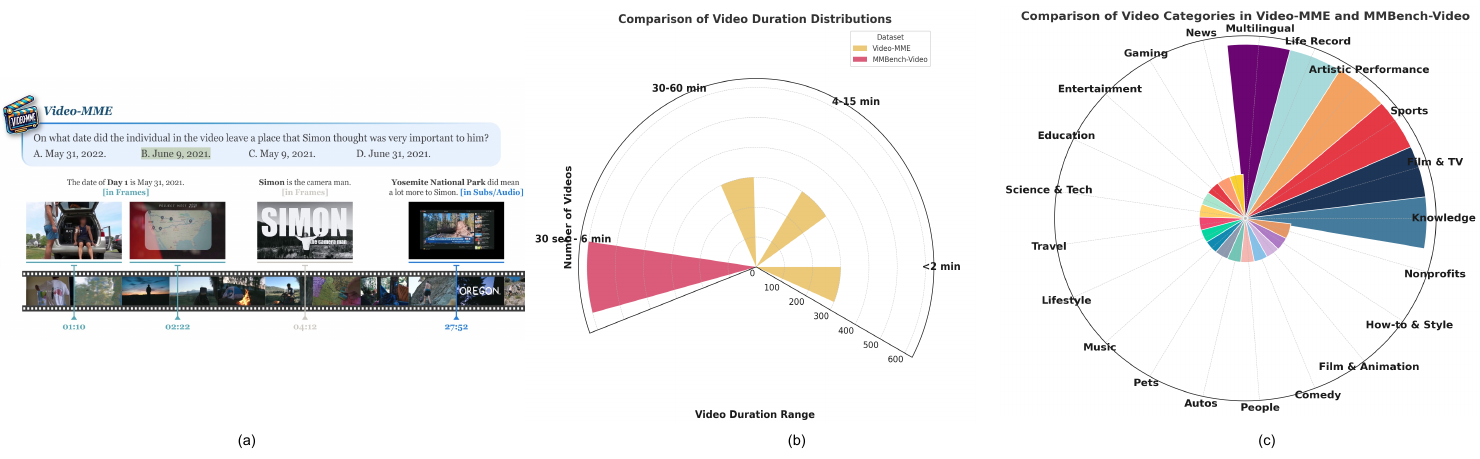}  
    \end{minipage}%
    \hfill
    \vspace{-10pt}
    \caption{Comparison of the Video-MME \citep{fu2024video} and MMBench-Video datasets \citep{MMBench}  in terms of video categories and duration distributions. The Video-MME dataset consists of 900 videos spanning six primary visual domains with 30 subfields, categorized into 300 short-term (<2 min), 300 medium-term (4-15 min), and 300 long-term (30-60 min) videos. In contrast, the MMBench-Video dataset comprises approximately 609 videos across 16 major categories, with durations ranging all from 30 seconds to 6 minutes. }
    \label{fig: BenchmarkData}
\end{figure*}

\begin{table*}[h]
    \centering
    \caption{Video-MME \citep{fu2024video} raw dataset structure, Q/A details in Json format.}
    \renewcommand{\arraystretch}{1.2}
    \begin{tabular}{|l|p{10cm}|}
        \toprule
        \rowcolor{gray!30} \textbf{Field} & \textbf{Details} \\
        \midrule
        Video ID & 001 \\
        Duration & Short \\
        Domain & Knowledge \\
        Sub-Category & Humanity \& History \\
        URL & \texttt{https://www.youtube.com/ watch?v=fFjy93ACGo8} \\
        VideoID & fFjy93ACGo8 \\
        Question ID & 001-2 \\
        Task Type & Information Synopsis \\
        Question & What is the genre of this video? \\
        \midrule
        \rowcolor{gray!30} \multicolumn{2}{|c|}{\textbf{Options}} \\
        \midrule
        A & It is a news report that introduces the history behind Christmas decorations. \\
        B & It is a documentary on the evolution of Christmas holiday recipes. \\
        C & It is a travel vlog exploring Christmas markets around the world. \\
        D & It is a tutorial on DIY Christmas ornament crafting. \\
        \midrule
        \textbf{Answer} & \textbf{A} \\
        \bottomrule
    \end{tabular} \label{table: videomme rawdata}
\end{table*}

\begin{table*}[h]
    \centering
    \caption{Comparison of Video Summarization: Gemini-2-Flash vs Qwen-7B. Here \emph{\textcolor{red!100}{the timestamps are all wrong}}.}
    \renewcommand{\arraystretch}{1.2}
    \begin{tabular}{|l|p{6cm}|p{6cm}|}
        \toprule
        \rowcolor{gray!30} \textbf{Category} & \textbf{Gemini-2-Flash} & \textbf{Qwen-7B} \\
        \toprule
        \textbf{Video ID} & P69idA8JO98 & P69idA8JO98 \\
        \textbf{Duration} & Long & Long \\
        \textbf{Domain} & Artistic Performance & Artistic Performance \\
        \textbf{Summary} & A stage performance of \textit{Snow White}. The Evil Queen consults the Magic Mirror, instructs Snow White to clean the castle, and the story unfolds as Snow White meets the Seven Dwarfs, receives the poisoned apple, collapses, and is revived by the Prince. & A fairy tale performance, likely \textit{Snow White and the Seven Dwarfs}. The video introduces characters, a forest scene, a confrontation between a queen and a prince, interactions between Snow White and the dwarfs, and ends with a song. \\
        \midrule
    \end{tabular}
    
    % Key Frames Section
    \vspace{0.5cm}
    \begin{tabular}{|l|p{12cm}|}
        \toprule
        \multicolumn{2}{|c|}{\textbf{Key Frames (Gemini-2-Flash)}} \\
        \toprule
        \textbf{Time} & \textbf{Description} \\
        \midrule
        \emph{\textcolor{red!100}{00:08}} & Magic Mirror reveals an angry face \\
        \emph{\textcolor{red!100}{00:42}} & Snow White in rags looking at her stepmother \\
        \emph{\textcolor{red!100}{00:46}} & The Evil Queen on a castle balcony \\
        \emph{\textcolor{red!100}{01:27}} & Dopey Dwarf dancing in silk costume \\
        \emph{\textcolor{red!100}{01:37}} & Ethereal dancer twirling with a deer \\
        \emph{\textcolor{red!100}{02:17}} & Snow White with basket approaching animals \\
        \emph{\textcolor{red!100}{02:57}} & Snow White collapses onto a stage of rocks \\
        \emph{\textcolor{red!100}{03:42}} & Snow White at a wishing well \\
        \emph{\textcolor{red!100}{05:09}} & The Evil Queen on a balcony speaking to a soldier \\
        \emph{\textcolor{red!100}{05:37}} & Snow White dancing in her new dress \\
        \emph{\textcolor{red!100}{06:09}} & Snow White and her prince hold hands \\
        \emph{\textcolor{red!100}{07:02}} & Snow White falls, animals mourn her \\
        \emph{\textcolor{red!100}{07:26}} & The Prince awakens Snow White with a kiss \\
        \emph{\textcolor{red!100}{08:00}} & Snow White is held up for celebration \\
        \emph{\textcolor{red!100}{08:07}} & Evil Queen standing on castle balcony \\
        \emph{\textcolor{red!100}{09:05}} & Snow White lies in a glass coffin as prince kneels \\
        \bottomrule
    \end{tabular}
    
    \vspace{0.5cm}
    \begin{tabular}{|l|p{12cm}|}
        \toprule
        \multicolumn{2}{|c|}{\textbf{Key Frames (Qwen-7B)}} \\
        \toprule
        \textbf{Time} & \textbf{Description} \\
        \midrule
        \emph{\textcolor{red!100}{00:00}} & Introduction of characters and setting \\
        \emph{\textcolor{red!100}{02:00}} & Scene with group of people in a forest \\
        \emph{\textcolor{red!100}{04:00}} & Confrontation between a queen and a prince \\
        \emph{\textcolor{red!100}{06:00}} & Introduction of the dwarfs as Snow White's friends \\
        \emph{\textcolor{red!100}{08:00}} & Scenes of the dwarfs working and interacting with Snow White \\
        \emph{\textcolor{red!100}{10:00}} & Snow White singing a song with the dwarfs \\
        \bottomrule
    \end{tabular}\label{table: timeacc}
    \label{table: outputresults}
\end{table*}

\begin{table*}
     \centering
    \caption{With and Without ALM for Summary and Key Frame: Using ALM for voiceover transcription improves the relevance and precision of the summary content \emph{\textcolor{blue!100}{(see the blue marked texts)}}. However, timing remains a significant issue.}
    \renewcommand{\arraystretch}{1.1}
    \begin{tabular}{p{2cm}|p{13cm}}
    \toprule
    \textbf{Component} & \textbf{Output Text} \\
    \midrule
    \textbf{Prompt (Qwen Model)}  &
    "Focusing on the content and workflow rather than specific logos or the color of text? After summarizing, list the key frames with brief captions in the format (00:00, caption). Ensure the analysis is accurate and avoid including any assumptions or extrapolations. Use a pharmaceutical expert domain perspective to enhance relevance and precision." \\
    \midrule
    \textbf{Whisper Model (Voice-over Transcription)}  &
    " "id": 0, "start": 7.72, "end": 13.6, "text": " PCR of course refers to pathological complete response where once the patient has surgery" ...\newline "\textbf{Full Transcription}": " PCR of course refers to pathological complete response where once the patient has surgery the pathologist does not find any cancer at all and pleasingly over the last sort of 15-20 years we've seen improvements in systemic treatment to such an extent that certainly for HER2 positive breast cancers we are now able to expect 50-60\% of patients who have a PCR following the neoadjuvant treatment and we know that the patients who have a PCR are the ones who are likely to do the best in the long term. " \\
    \midrule
    \textbf{Qwen Model Output (Before Transcription)} &
    \textbf{VLM Output}: "The video appears to be a medical demonstration, likely related to the Roche brand, which is known for its involvement in healthcare and diagnostics. The video showcases a 3D model of a human body with various organs highlighted, including the liver and brain. The model is displayed within a transparent box, possibly representing a medical device or a simulation environment. \newline
    \textbf{Key Frames with Captions}: 00:00 - Introduction: The video begins with a close-up of a transparent box containing a 3D model of a human body. 01:00 - Organ Highlighting: The liver and brain are highlighted in the model, indicating a focus on these organs. 02:00 - Interactive Display: The model rotates, providing a better view of the highlighted organs. 03:00 - Data Overlay: Digital text and graphics appear, likely representing medical data or information related to the highlighted organs. 04:00 - Conclusion: The video ends with a summary or conclusion, possibly emphasizing the importance of the highlighted organs or the role of the device in medical diagnostics. The video likely aims to educate viewers about the importance of the liver and brain in the human body and how the device or simulation can be used to study or diagnose related conditions."
    " \\
    \midrule
    \textbf{Qwen Model Output (After Transcription Integration)}  &
     \textbf{VLM Output}: "Here is a summary focusing on the content and workflow from a pharmaceutical expert's perspective, along with key frames and brief captions" 
    Summary: The video provides an explanation of the term "\emph{\textcolor{blue!100}{pathological complete response" (pCR)}} and its significance in cancer treatment, particularly for \emph{\textcolor{blue!100}{HER2-positive breast cancer}}. The speaker, a medical professional, \emph{\textcolor{blue!100}{defines pCR as the absence of any detectable cancer cells in the surgical specimen after neoadjuvant (pre-operative) systemic treatment}}. He highlights the remarkable progress made in systemic therapies over the past 15-20 years, \emph{\textcolor{blue!100}{enabling 50-60\% of HER2-positive breast cancer patients to achieve pCR following neoadjuvant treatment}}. The speaker \emph{\textcolor{blue!100}{emphasizes that patients who achieve pCR have the best long-term outcomes}}, underscoring the importance of this endpoint in cancer management. 
    \textbf{Key Frames}:(00:07) Introduction to the concept of pathological complete response (pCR).(00:19) Progress in systemic treatments improving pCR rates. (00:24) Specifically mentioning 50-60\% pCR rates in HER2-positive breast cancer.(00:33) Linking pCR to better long-term outcomes for patients." 
 \\
    
    \bottomrule
    \end{tabular} \label{table: sampleoutput}
\end{table*}

\begin{figure*}[h]
    \centering
    \begin{minipage}{1\textwidth}
        \centering
        \includegraphics[width=1\linewidth]{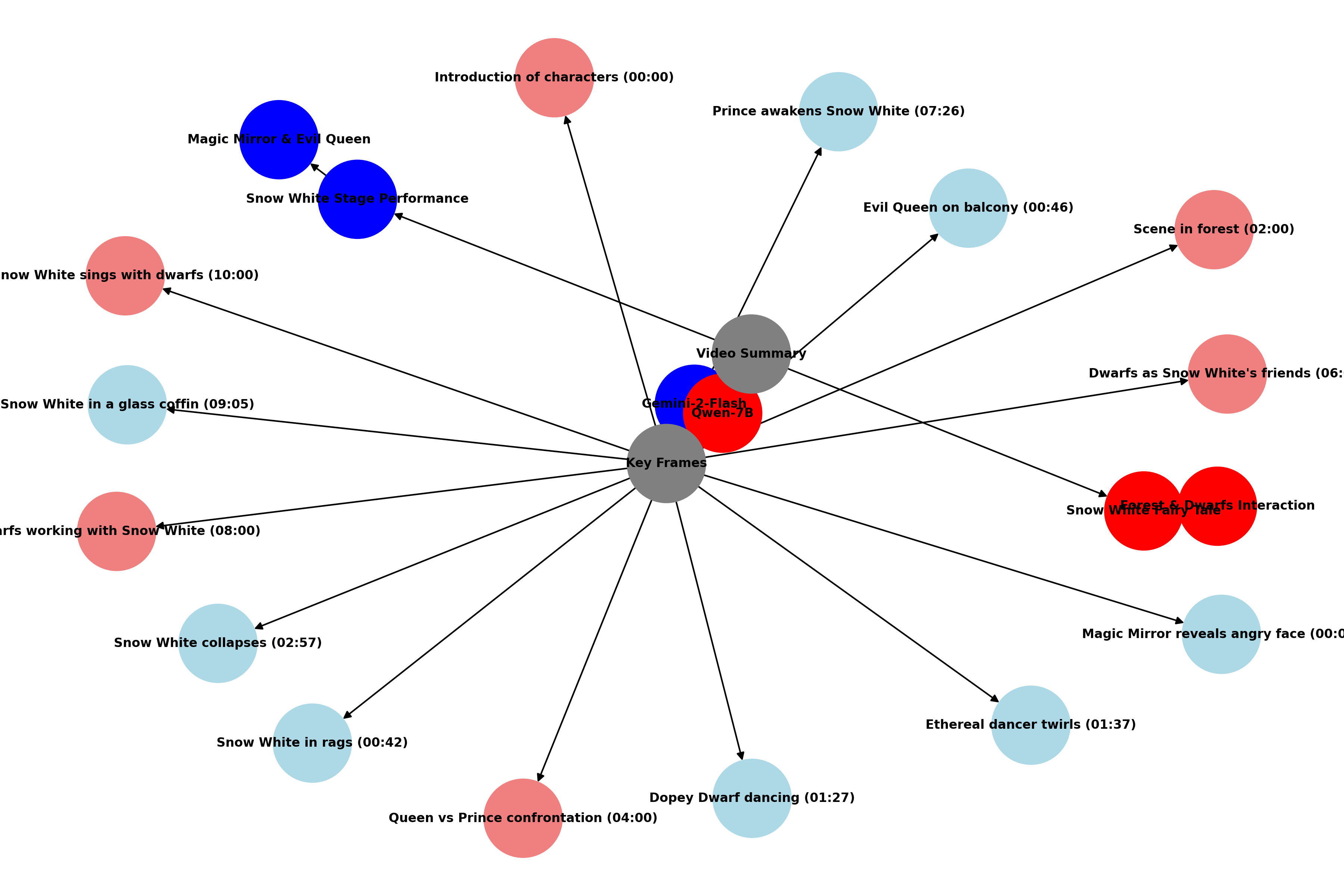}  
    \end{minipage}%
    \hfill
    \vspace{-5pt}
    \caption{Knowledge graph for summary and key frames using Qwen and Gemini models. The knowledge graph visualizes the comparison between Gemini-2-Flash and Qwen-7B in summarizing a 'Snow White' stage performance. Each key frame from Gemini-2 Flash is marked in light blue and video summary in dark blue, while Qwen-7B's key frames are in light red, and video summary in red. The central node represents the key frames, with connections showing their relationships to each model's summary. Gemini-2-Flash emphasizes narrative elements such as the Magic Mirror, the Evil Queen, and the climax involving Snow White's revival, while Qwen-7B structures the story around broad thematic transitions like character introductions, forest scenes, and musical elements. This graph presents a structured comparison of the keyframes extracted by Gemini-2-Flash and Qwen-7B from a 'Snow White' performance. \emph{\textcolor{red!100}{The blue nodes represent Gemini-2-Flash’s emphasis on theatrical storytelling, focusing on individual character moments, while the red nodes highlight Qwen-7B’s broader narrative structure}}, including interactions between Snow White and supporting characters. Additionally, the red nodes are more widely distributed, whereas the blue nodes are clustered more closely, indicating a difference in granularity and focus.}
    \label{fig: knowledgenode}
\end{figure*}

\begin{table*}[h]
    \centering
    \caption{Ablation Study on FPS (1): Attention Mechanism Dependence on FPS. Comparison of Speed, Accuracy, and Completeness Across Leading Attention Mechanisms on Short Videos (<120s). \emph{\textcolor{red!100}{This study highlights the strong dependence of FPS on each model's performance. For short videos, FlashAttention is recommended over SDPA}}. *Default settings as per the Video-MME benchmark (FPS = 1), with no additional audio ALM transcription fed into the Qwen model.}
    \renewcommand{\arraystretch}{1.2} % Adjust row spacing
    \setlength{\tabcolsep}{6pt} % Adjust column spacing
    \small % Reduce font size
    \begin{tabular}{c|c|c|c}
        \toprule
        \rowcolor{gray!20} \textbf{Experiments on \textcolor{red}{Short Videos*}} & \textbf{Processing Time} & \textbf{Total Answered (\%)} & \textbf{Correct Answered (\%)} \\
        \midrule
        SDPA (1 FPS) &  44m 12s & 6\% & 64.81\% \\
        FlashAttention (1 FPS) & 1h 30m 12s & 100\% & \textcolor{red}{70.78\%} \\
        \bottomrule 
        
        \toprule
        \rowcolor{gray!20} \textbf{Experiments on \textcolor{red}{All Length Videos}} & \textbf{Processing Time} & \textbf{Total Answered (\%)} & \textbf{Correct Answered (\%)} \\
        \midrule
        SDPA (0.1 FPS) & 4h 37m 2s & 37\% & \textcolor{red}{58.73\%} \\
        FlashAttention (0.1 FPS) & 7h 40m 17s & 100\% & 54.81\% \\
        SDPA (0.01 FPS) & 2h 12m 37s & 87\% & 48.40\% \\
        FlashAttention (0.01 FPS) & 1h 50m 11s & 100\% & 48.55\% \\
        \bottomrule
    \end{tabular}
    \label{table:fps}
\end{table*}

\begin{table*}[t]
    \centering
    \caption{Ablation Study on FPS (2) \emph{\textcolor{red!100}{Reducing FPS does not necessarily help the Qwen model answer more questions correctly}}. In fact, it can have a negative impact, as lower frames per second lead to missing information. Here, the completeness percentage increases from 37\% to 87\% significantly, but the accuracy drop from 58.73\% to 48.40\%. However, \emph{\textcolor{red!100}{with the support of audio ALM transcription, accuracy is maintained}}, improving from 58.73\% to 61.80\% when FPS = 0.1, and from 48.40\% to 49.36\%. This further validates our first finding from a different perspective.}
    \renewcommand{\arraystretch}{1.2}
    \begin{tabular}{c|c|c}
        
        \toprule
        \rowcolor{gray!20} \textbf{Experiments on  All Length Videos}  & \textbf{Total Answered (\%)} & \textbf{Correct Answered (\%)} \\
        \midrule
         SDPA (0.1 FPS) without Audio Transcription & 37\% & 58.73\%\\
         SDPA (0.1 FPS) with Audio Transcription  & 26.78\% & \textcolor{red!100}{61.80\%} \\
         SDPA (0.01 FPS) without Audio Transcription & 87\% & 48.40\% \\
         SDPA (0.01 FPS) with Audio Transcription  & 68.92\% & 49.36\% \\
        \bottomrule
    \end{tabular}\label{table: fps2}
\end{table*}

\begin{figure*}[t]
    \centering
    \begin{minipage}{1\textwidth}
        \centering
        \includegraphics[width=1\textwidth]{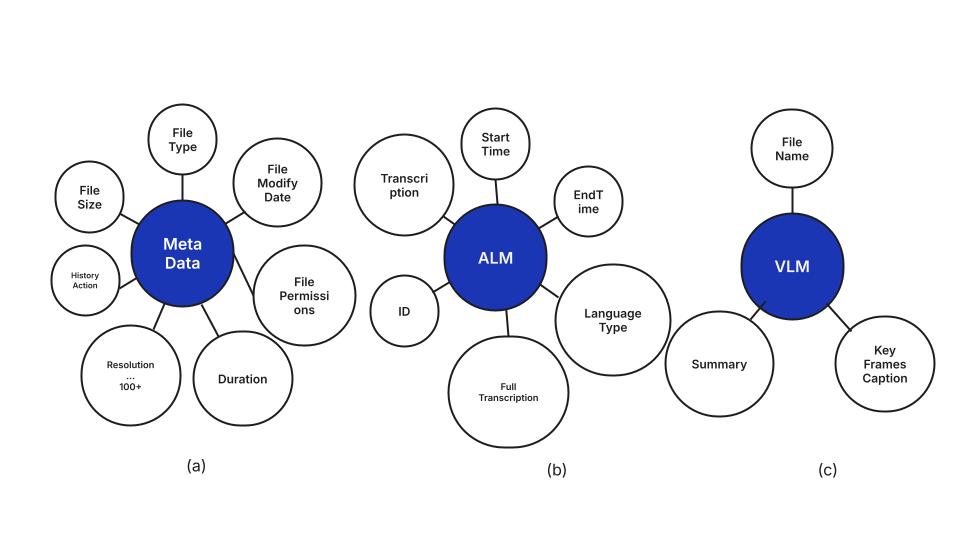} 
    \end{minipage}%
    \vspace{-10pt}
    \caption{Deliverable Attributes. Define each GenAI output file attributes. *ALM stands for Audio Large Language model generated output, and *VLM represents Video Large Language model generated output.}
    \label{fig: deliverable}
\end{figure*}

\begin{table*}[h]
    \centering
    \caption{Distribution of Property Audio and Video Data Across Medical Diseases Specialties.}
    \renewcommand{\arraystretch}{1.2}
    \rowcolors{2}{gray!15}{white}
    \begin{tabular}{l|c|c||l|c|c}
        \toprule
        \textbf{Specialty} & \textbf{Audio} & \textbf{Video} & \textbf{Specialty} & \textbf{Audio} & \textbf{Video} \\
        \midrule
        Oncology & 208 & 8934 & Ophthalmology & 159 & 2862 \\
        Cardiovascular & 1 & 14 & Respiratory Disease & 16 & 467 \\
        Dermatology & 0 & 30 & Nephrology & 1 & 380 \\
        Hematology & 67 & 3606 & Not Applicable & 59 & 2853 \\
        Immunology & 144 & 510 & Movement Disorder & 9 & 289 \\
        Infectious Disease & 1 & 239 & Inflammatory Disease & 20 & 222 \\
        Metabolism & 0 & 6 & Neuroscience & 202 & 4914 \\
        \bottomrule
    \end{tabular}
    \label{table:roche-data}
\end{table*}

\begin{table*}[h]
\centering
\caption{The backbones, hyper-parameters, and prompt settings of \textbf{selective} SOTA VLMs.}
\resizebox{\textwidth}{!}{
\begin{tabular}{@{}p{4cm}|p{1cm}|p{2cm}|p{2cm}|p{3cm}|p{5cm}@{}}
\toprule
\multicolumn{1}{c|}{Model Description} & Type & Token Limit & API Price in Dollars & \multicolumn{1}{c|}{Modality}& \multicolumn{1}{c}{Prompt Example} \\ 
\midrule

GPT-4 Turbo, The latest GPT-4 model with improved instruction, reproducible outputs, parallel function calling. Returns max of 4,096 output tokens. Training data up to Apr 2023 & gpt-4-1106-preview &  Input token limit:128$K$ & Input $0.06/K$ Tokens. Output $0.12/K$ Tokens & Text Generation, Chat Completion, Image to Text & Could you please provide a summary of this video based on sample frames focusing on the content and workflow rather than specific logos or the color of text? After summarizing, list the key frames with brief captions in the format (00:00, caption). Ensure the analysis is accurate and avoid including any assumptions or extrapolations. Use an expert domain perspective to enhance relevance and precision. Do not repeat sentences or focus on QR codes or logos. \\                                                                             
\hline
Qwen2-VL. updated on Huggingface Jan. 2025 & Qwen2-VL-7B & Input token limit:$32K$ &  Opensource & Text Generation, Video to Text  &  Could you please provide a summary of this video, focusing on the content and workflow rather than specific logos or the color of text? After summarizing, list the key frames with brief captions in the format (00:00, caption). Ensure the analysis is accurate and avoid including any assumptions or extrapolations. Use an expert domain perspective to enhance relevance and precision. Do not repeat sentences or focus on QR codes or logos.\\
\hline
Gemini 2.0 Flash Model, released data 05th Feb 2025 & Gemini 2.0 Flash 001 & Input token limit:$1048K$
Output token limit: $8K$ & Per 1M tokens in USD: API Cost 0.10 (text / image / video)
0.70 (audio). Output 0.40. e.g., 300 Long Videos from Video-MME costed 56.62 & Audio, images, video, text, and PDF to text & Could you please provide a summary of this video, focusing on the content and workflow rather than specific logos or the color of text? After summarizing, list the key frames with brief captions in the format (00:00, caption). Ensure the analysis is accurate and avoid including any assumptions or extrapolations. Use an expert domain perspective to enhance relevance and precision. Do not repeat sentences or focus on QR codes or logos.\\ 
\bottomrule
\end{tabular}
}
\label{tab:hypo}
\end{table*}

\begin{table*}
     \centering
    \caption{Area of Impact 1: AI-driven Video Processing for Medical Diagnosis.}
    \renewcommand{\arraystretch}{1.2}
    \begin{tabular}{p{3cm}|p{12cm}}
    \toprule
    \textbf{Category} & \textbf{Applications} \\
    \midrule
    \textbf{Radiology and Imaging} & 
    \begin{itemize}
    \item \textbf{MRI/CT Scan Video Processing}: Advanced AI can analyze full-length MRI or CT scans in motion (e.g., cardiac MRI or functional MRI), detecting anomalies faster than manual review.
    \item \textbf{Ultrasound Interpretation}: AI-powered real-time video analysis can help with fetal health assessments, echocardiography, and liver disease detection.
\end{itemize} \\
\midrule
\textbf{Endoscopy and Surgery} & 
\begin{itemize}
    \item \textbf{Colonoscopy Polyp Detection}: AI can process hours of colonoscopy footage to detect polyps in real-time, improving colorectal cancer screening.
    \item \textbf{Robotic Surgery Assistance}: AI-driven video processing can provide real-time insights to surgeons, flagging anomalies or suggesting procedural adjustments.

\end{itemize} \\
\midrule
\textbf{Neurology and Movement Disorders} & 
\begin{itemize}
    \item \textbf{Seizure and Tremor Analysis}: AI can analyze EEG-video recordings to classify epilepsy types.
    \item \textbf{Parkinson’s and ALS Monitoring}: AI can assess gait, facial expressions, and movement from patient videos for early diagnosis and tracking progression.

\end{itemize} \\
\midrule
\textbf{Ophthalmology and Dermatology} & 
\begin{itemize}
    \item \textbf{Retinal Scan Analysis}: AI models can process retinal scan videos to detect early diabetic retinopathy or macular degeneration.

    \item \textbf{Skin Cancer Detection}: Dermatologists can use AI-enhanced dermoscopy video processing to detect melanoma more accurately.
\end{itemize} \\
\midrule
\textbf{Pathology and Microscopy} & 
\begin{itemize}
    \item AI can analyze continuous microscopy footage to identify cancerous cells, bacterial infections, or rare hematological conditions in blood samples faster than human pathologists.
\end{itemize} \\
\midrule
\textbf{Remote Patient Monitoring} & 
\begin{itemize}
    \item Wearable devices that record and process patient videos (e.g., heart rate monitors, movement trackers) can enable early diagnosis of conditions like heart arrhythmias or sleep apnea at lower costs.
\end{itemize} \\
\bottomrule
\end{tabular}
\label{tab:business1}
\end{table*}

\vspace{1cm}

\begin{table*}
\centering
    \caption{Area of Impact 2: AI-driven Video Processing for Drone/Satellite Data Analysis.}
    \renewcommand{\arraystretch}{1.2}
    \begin{tabular}{p{3cm}|p{12cm}}
    \toprule
    \textbf{Category} & \textbf{Applications} \\
    \midrule
\textbf{Agriculture and Environment} & 
\begin{itemize}
    \item \textbf{Crop Health Monitoring}: AI-driven video analysis can quickly identify stressed crops, pest infestations, or nutrient deficiencies.

    \item \textbf{Deforestation and Land Use}: Detecting illegal logging or monitoring ecosystem changes becomes faster and cheaper.
    \item \textbf{Disaster Assessment}: Rapid damage assessment after hurricanes, earthquakes, or floods helps authorities respond effectively.

\end{itemize} \\
\midrule
\textbf{Infrastructure and Urban Planning} & 
\begin{itemize}
    \item \textbf{Road and Bridge Inspections}: AI can process high-resolution drone footage to detect cracks, erosion, or weak points.
    \item \textbf{Traffic and Urban Planning}: Satellite video can track congestion patterns and optimize urban development.
\end{itemize} \\
\midrule
\textbf{Defense and Security} & 
\begin{itemize}
    \item \textbf{Surveillance and Threat Detection}: Automated analysis of drone/satellite feeds can detect anomalies, unauthorized activities, or suspicious movements.

    \item \textbf{Border and Maritime Security}: Continuous video monitoring can identify smuggling, illegal crossings, or unauthorized vessel movements.

\end{itemize} \\
\midrule
\textbf{Healthcare and Epidemiology} & 
\begin{itemize}
    \item \textbf{Mosquito-Borne Disease Prevention}: Satellite video can help detect standing water bodies where mosquitoes breed, aiding in malaria/dengue prevention.

    \item \textbf{Air Pollution and Public Health}: Fast video analysis can track pollution hotspots, correlating air quality data with disease outbreaks.

\end{itemize} \\
\bottomrule
\end{tabular}
\label{tab:business2}
\end{table*}

\begin{figure*}[t]
    \centering
    \begin{minipage}{1\textwidth}
        \centering
        \includegraphics[width=1\textwidth]{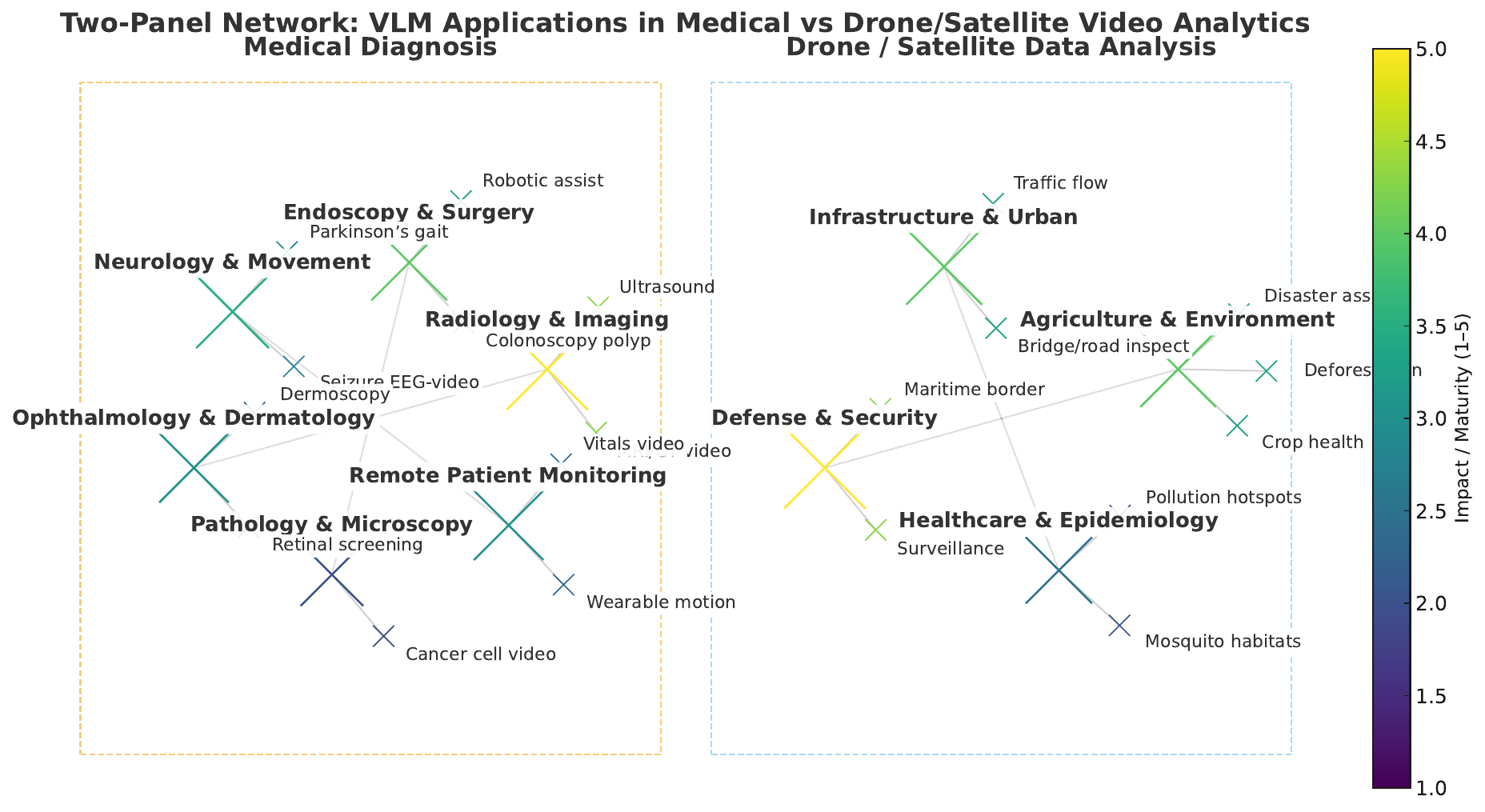} 
    \end{minipage}%
    \vspace{-10pt}
    \caption{\textbf{Two-Panel Network of Vision--Language Model (VLM) / AI Applications Across Medical and Drone/Satellite Video Analytics.}
Node color and size encode an \emph{impact/maturity score} (1--5).
Impact is estimated as a composite index: \textbf{$\mathrm{Impact}=0.4R+0.2D+0.2A+0.2C$}, where
$R$ denotes normalized research intensity (publications indexed in PubMed, IEEE Xplore, Scopus from 2020--2025 containing ``video + AI or VLM + domain''),
$D$ dataset availability (standardized/public video datasets),
$A$ application readiness (evidence of clinical or industrial deployment and/or regulatory signals),
and $C$ cross-domain generalizability (extent of transferability to other domains or modalities).
Scores are normalized to $[1,5]$ (5 = high maturity, well-validated, standardized datasets, active commercial use) and were assigned approximately as:
Radiology \& Imaging 5.0 with extensive literature (>200 papers, 2020–2025), benchmark datasets (e.g., MIMIC-CXR, cardiac MRI) \cite{Li2021MedicalDatasets,Najjar2023RadiologyReview, PintoCoelho2023AIImaging}, and multiple FDA-approved AI tools \cite{FDA_AIML_List,ACR_AICentral}; Endoscopy \& Surgery 4.0 with validated prototypes for polyp detection and robotic surgery support\cite{Saeidi2022STAR}, but moderate dataset availability \cite{Borgli2020HyperKvasir,Ali2023PolypGen,Rios2023Cholec80CVS}; Neurology \& Movement 3.5; Ophthalmology \& Dermatology 3.0 with strong still-image AI base (fundus/OCT in ophthalmology; dermoscopy in dermatology), including FDA-cleared autonomous DR screening and large image datasets/challenges \cite{FDADEN180001,Oshika2024AIOph, Xu2024DRReview,ISICArchive,Yilmaz2024DERM12345,Behara2024DermAI} but few real-time video pipelines available \cite{Mahmoud2024CatStep,Ghamsarian2024Cataract1K,BJSOpen2024SurgVideo} ;
Pathology \& Microscopy 2.0 with only limited continuous video/time-lapse microscopy datasets and research mostly whole-slide image (WSI)–based \cite{Fatima2024DigPathReview,TanGarcia2025CompPath}; Remote Patient Monitoring 3.0 with active wearable and webcam-based monitoring studies and few open datasets \cite{Haque2025rPPGBias,Huang2023npjDM} ; Agriculture \& Environment 4.0 with large drone video datasets and real-world agricultural monitoring systems and high industrial uptake \cite{Strzepek2023AgricultureUAV, Wang2025UAVSurvey};
Infrastructure \& Urban 4.0 with drone/traffic inspection systems widely deployed and mature technical readiness \cite{Du2018UAVDT}; Defense \& Security 5.0 with advanced object tracking and anomaly detection systems in full industrial operation \cite{Zhang2022ByteTrack} ; Healthcare \& Epidemiology 2.5 with exploratory studies linking environmental video data with disease risk or pollution metrics \cite{Wei2023GlobalPM25}.
Smaller satellite nodes indicate representative sub-tasks within each category.}
    \label{fig: business3}
\end{figure*}

\clearpage

\end{document}